\documentclass[manuscript, nonacm]{jair}

\setcopyright{cc}
\copyrightyear{2025}
\acmDOI{10.1007/s40747-026-02251-1}

\JAIRAE{}
\JAIRTrack{} 
\acmMonth{3}
\acmYear{2026}
\usepackage{algorithm}%
\usepackage{algorithmic}%
\usepackage{multirow}
\usepackage{threeparttable}
\usepackage{booktabs}
\usepackage{graphicx}
\usepackage{subcaption}
\usepackage{amsmath}
\DeclareMathOperator{\E}{\mathbb{E}}
\newcommand{\independent}{\mathrel{\perp\mspace{-10mu}\perp}}

\RequirePackage[
   datamodel=acmdatamodel,
   style=acmnumeric,
   backend=biber,
   giveninits=true
   ]{biblatex}

\begin{document}

\title[Dynamic Uncertainty Aware Divide2Conquer]{DUA-D2C: Dynamic Uncertainty Aware Method for Overfitting Remediation in Deep Learning}

\author{Md. Saiful Bari Siddiqui}
\authornote{Corresponding Author.}
\orcid{0009-0000-7781-0966}
\email{saiful.bari@bracu.ac.bd}
\affiliation{%
  \institution{BRAC University}
  \city{Dhaka}
  \country{Bangladesh}
}

\author{Md Mohaiminul Islam}
\orcid{0009-0003-5716-4946}
\email{mohaiminul@cse.uiu.ac.bd}
\affiliation{%
  \institution{United International University}
  \city{Dhaka}
  \country{Bangladesh}}

\author{Md. Golam Rabiul Alam}
\orcid{0000-0002-9054-7557}
\email{rabiul.alam@bracu.ac.bd}
\affiliation{%
  \institution{BRAC University}
  \city{Dhaka}
  \country{Bangladesh}
}

\renewcommand{\shortauthors}{Siddiqui, Islam \& Alam}

\begin{abstract}
{\bf Background:} 
    Overfitting remains a significant challenge in deep learning, often arising from data outliers, noise, and limited training data. To address this, we previously proposed the Divide2Conquer (D2C) method, which partitions training data into multiple subsets and trains identical models independently on each. This strategy enables learning more consistent patterns while minimizing the influence of individual outliers and noise.  
    
    {\bf Objectives:}
    D2C's standard aggregation typically treats all subset models equally or based on fixed heuristics (like data size), potentially underutilizing information about their varying generalization capabilities. Building upon this foundation, we introduce Dynamic Uncertainty-Aware Divide2Conquer (DUA-D2C), an advanced technique that refines the aggregation process. 
    
    {\bf Methods:}
    DUA-D2C dynamically weights the contributions of subset models based on their performance on a shared validation set, employing a novel composite score of accuracy and normalized prediction entropy. This intelligent aggregation allows the central model to preferentially learn from subsets yielding more generalizable and confident edge models, thereby more effectively combating overfitting.  
    
    {\bf Results:}
    In this work, we provide a rigorous theoretical justification for this approach, analytically demonstrating how dynamic parameter fusion reduces model variance. Empirical evaluations on benchmark datasets spanning image, audio, and text domains demonstrate that DUA-D2C significantly improves generalization. Our analysis includes evaluations of decision boundaries, loss curves, and ablation studies, highlighting that DUA-D2C provides additive performance gains even when applied on top of standard regularizers like Dropout. 
    
    {\bf Conclusions:}
    This study establishes it as a theoretically grounded and effective approach to combating overfitting in deep learning. The source codes for this study are available on GitHub at \href{https://github.com/Saiful185/DUAD2C}{https://github.com/Saiful185/DUAD2C}.
\end{abstract}




\maketitle

\section{Introduction}\label{intro}

Deep Learning models often achieve high performance on their training data, yet this performance can drop significantly on unseen data from different distributions, a phenomenon known as overfitting. Numerous methods have been proposed over the years to mitigate overfitting \cite{ying2019overview}. Early Stopping \cite{morgan1989generalization}, while intuitive, may limit a model's learning potential \cite{prechelt2002early}. Network reduction simplifies models to reduce variance but can restrict the learning of complex features \cite{furnkranz1997pruning}. Data augmentation is a common approach \cite{sun2014deep}, but selecting suitable techniques can be challenging, and acquiring additional data often requires substantial effort. Regularization, particularly methods like Dropout \cite{srivastava2014dropout} which randomly deactivates connections, is widely used. However, even Dropout can sometimes degrade performance due to issues like internal covariance shifts \cite{li2019understanding}. A common thread is that despite these techniques, the model ultimately accesses the entire training distribution simultaneously, allowing robust neural networks to potentially learn complex, noise-dependent correlations across the full dataset.

This observation motivates our work. We posit that preventing the neural network from training on the entire dataset simultaneously can be beneficial. Intuitively, we should combine multiple models that train on different portions of the data so that the model cannot familiarize itself with all of it at once. Instead, by combining multiple models trained on isolated portions of the data, a consensus can be achieved that focuses on general patterns while ignoring subset-specific noise. \textbf{Crucially, this represents a paradigm shift in how data partitioning is utilized.} In frameworks like Federated Learning (FL), data partitioning is a constraint imposed by privacy; here, we treat partitioning as a deliberate \textbf{``architectural regularization'' strategy}. By forcing models to learn from limited views, we prevent any single model from memorizing the global noise floor.

To this end, the Divide2Conquer (D2C) framework \cite{siddiqui2024d2c} was recently introduced. D2C partitions training data into multiple subsets and trains identical models independently on each. After a fixed number of local epochs, the parameters from these \textbf{edge models} (models trained on subsets) are aggregated through weighted averaging. While the original D2C work provided empirical evidence for this approach, it lacked a rigorous theoretical basis for why partitioning reduces variance in deep learning. Furthermore, its standard aggregation mechanism treats all edge models equally (or based solely on data size), potentially underutilizing information about their varying generalization capabilities.

Building upon this foundation, we introduce \textbf{Dynamic Uncertainty-Aware Divide2Conquer (DUA-D2C)}, an advanced technique that perfects the aggregation process. Unlike standard Federated Averaging (FedAvg), which aggregates based on data \textit{quantity} ($n_k/N$), thereby risking the integration of noise from corrupted data shards, DUA-D2C introduces a \textbf{quality-aware aggregation}. We dynamically weight the contributions of edge models based on a composite score of their validation accuracy and \textbf{normalized prediction entropy}. This intelligent, adaptive weighting scheme acts as a filter, allowing the central model to preferentially learn from edge models that demonstrate superior generalization and confidence. This effectively suppresses \textit{confidently wrong} or overfitted models before they corrupt the central parameters.

Our choice of \textbf{entropy} for uncertainty quantification is uniquely tailored to this centralized, iterative framework. Unlike Bayesian approximation methods (e.g., MC Dropout~\cite{gal2016dropoutbayesianapproximationrepresenting}), which require multiple forward passes per sample, incurring prohibitive computational costs during iterative aggregation, entropy provides a distribution-aware uncertainty signal in a single forward pass. Because DUA-D2C operates in a centralized regime where a reliable validation set is available, entropy serves as an efficient and sufficient proxy for model confusion. This allows us to perform high-frequency, quality-based filtering of edge models with negligible computational overhead, a capability that is difficult to achieve in standard federated or ensemble settings.

It is vital to distinguish DUA-D2C from existing federated methods. While our architecture borrows the iterative ``train-then-aggregate'' structure of Federated Optimization \cite{konevcny2015federated}, the objectives are fundamentally different. FL algorithms are designed to operate under strict constraints where data is inherently decentralized; their goal is to mitigate performance loss caused by non-IID data. DUA-D2C \textit{inverts} this logic, utilizing induced partitioning to \textit{exceed} centralized performance. Moreover, unlike advanced aggregation strategies such as \textbf{FedProx} \cite{li2020fedprox} or \textbf{FedAtt} \cite{ji2019learning}, which constrain local updates to prevent drift or weight models based on parameter distance, DUA-D2C imposes no such geometric constraints. We argue that parameter deviation is desirable if it stems from diverse feature learning. DUA-D2C resolves the ambiguity between ``good diversity'' and ``bad noise'' not by correcting optimization drift, but by filtering models based on explicit validation performance.

Furthermore, DUA-D2C differs fundamentally from Bagging \cite{ghojogh2019theory} or traditional Ensemble Learning in terms of aggregation mechanics and deployment efficiency. Standard ensembles typically maintain multiple models throughout the lifecycle and aggregate their \textit{predictions} (e.g., via soft voting) during inference. While effective for variance reduction, this approach incurs high computational overhead, requiring $N$ forward passes for every single prediction ($N \times$ inference cost). In contrast, DUA-D2C aggregates \textit{parameters} (weights) during the training phase. By fusing the ``wisdom'' of the subsets into a single, consolidated central model, DUA-D2C retains the generalization benefits of ensembling but incurs \textbf{zero additional inference overhead}, making it highly efficient for real-world deployment. Crucially, because DUA-D2C operates at the macroscopic data level, it is orthogonal to microscopic regularizers like Dropout or Batch Normalization. This allows it to be seamlessly stacked on top of existing architectures to provide additive gains in generalization without conflict.

We summarize our contributions through this study below:
\begin{itemize}
    \item \textbf{Formalization of 'Partitioning as Regularization':} We propose and validate a novel training paradigm where data partitioning is utilized not as a privacy constraint (as in Federated Learning), but as a deliberate \textit{structural regularizer}. We demonstrate that intentionally isolating training subsets prevents edge models from co-adapting to global noise and outliers, effectively repurposing decentralized architectures to solve the centralized problem of overfitting.
    \item \textbf{Dynamic Uncertainty-Aware D2C (DUA-D2C):} To perfect this paradigm, we introduce DUA-D2C, an advanced aggregation technique that overcomes the limitations of static averaging. DUA-D2C employs an intelligent, dynamic weighting scheme based on validation accuracy and \textit{normalized prediction entropy}, allowing the central model to actively filter out noise by preferentially learning from reliable, well-generalizing edge models.
    \item \textbf{Theoretical Justification:} We provide a rigorous mathematical framework supporting the D2C hypothesis. We analytically demonstrate how the ``divide and conquer'' strategy reduces model variance and extend this derivation to show how DUA-D2C's weighted aggregation further minimizes the theoretical error bound compared to simple averaging.
    \item \textbf{Extensive Empirical Validation:} We demonstrate through experiments on six diverse datasets (spanning image, audio, and text domains) that DUA-D2C consistently outperforms both traditional monolithic training and static D2C approaches. We further provide a comprehensive sensitivity analysis of the hyperparameters and discuss the trade-offs, establishing DUA-D2C as a robust solution for Big Data scenarios.
\end{itemize}

The paper is organized as follows. In Section \ref{related-work}, we discuss relevant literature and differentiate our work from FedAvg and Ensembling. We establish the theoretical justification behind the D2C hypothesis in Section \ref{theory}. In Section \ref{method}, we detail the DUA-D2C methodology and its aggregation algorithm, followed by the experimental specifications in Section \ref{exp}. In Section \ref{result}, we evaluate our approaches through empirical experiments and analyze the results. We discuss the limitations in Section \ref{lim}. Section \ref{conc} contains concluding remarks.

\section{Related Works} \label{related-work}

\subsection{Generalization and Regularization in Deep Learning}
Overfitting remains a fundamental challenge in deep learning, prompting extensive research into regularization techniques~\cite{Bejani2021OverfittingReview, Aliferis2024OverfittingReview}. Traditional methods focus on constraining model capacity or introducing stochasticity at the microscopic level. For instance, \textbf{Weight Decay} adds a penalty term to the loss function, while \textbf{Dropout}~\cite{srivastava2014dropout} prevents neuron co-adaptation by randomly deactivating units during training. \textbf{Batch Normalization}~\cite{ioffe2015batch} addresses internal covariate shift, offering implicit regularization benefits. Other approaches like DeCov~\cite{cogswell2015reducing} focus on decorrelating representations, and $L_{1/4}$ regularization~\cite{kolluri2020reducing} attempts to overcome limitations of L1/L2 norms.

\textbf{Ensemble Learning} takes a different approach by combining multiple hypotheses. Techniques like ELVD~\cite{shanmugavel2023novel} and parallel ensemble networks~\cite{kim2020ensemble} demonstrate that aggregating diverse models can significantly reduce variance. Similarly, uncertainty-guided ensembles~\cite{Zhang2024UncertaintyEnsemble} have shown promise in semi-supervised settings. However, standard ensembles typically aggregate \textit{predictions} (voting), which incurs a high computational cost ($N \times$ inference time) during deployment.

\textbf{Data Partitioning strategies}, such as K-Fold Cross-Validation~\cite{korjus2016efficient}, share conceptual similarities with our approach by creating subsets. However, K-Fold is primarily an evaluation metric where models eventually see the full dataset (across folds). In contrast, our approach utilizes partitioning as a strict \textbf{structural regularizer}, ensuring that no single model ever accesses the global dataset simultaneously, thereby preventing the memorization of global noise patterns.

\subsection{Aggregation Strategies: Federated and Centralized}
The mechanics of aggregating distributed models have been heavily explored in \textbf{Federated Learning (FL)}. The foundational \textit{FedAvg}~\cite{mcmahan2017fedavg} algorithm aggregates weights based on data quantity. Subsequent iterations like \textit{FedProx}~\cite{li2020fedprox} and \textit{SCAFFOLD}~\cite{karimireddy2020scaffold} introduce regularization terms to handle non-IID data drift, while \textit{FedNova}~\cite{wang2020fednova} normalizes local updates. However, these methods are designed to mitigate the \textit{downsides} of partitioning (privacy constraints), whereas we aim to exploit partitioning as a \textit{feature}.

In \textbf{Centralized Learning}, weight aggregation is increasingly recognized as a generalization tool. \textit{Stochastic Weight Averaging (SWA)}~\cite{izmailov2018swa} averages weights over training time to find flatter minima. \textit{Model Soups}~\cite{wortsman2022soups} demonstrates that averaging fine-tuned models in parameter space improves robustness without the inference cost of ensembles.

\subsection{Methodological Distinction and Unique Contributions}
\textit{Dynamic Uncertainty-Aware Divide2Conquer (DUA-D2C)} synthesizes concepts from federated and centralized learning into a unified framework. However, it is methodologically distinct from existing approaches in terms of its objective, aggregation logic, and deployment efficiency.

\subsubsection*{Contrast with Advanced Federated Aggregation}
While DUA-D2C adopts the iterative aggregation structure of FL, its mechanism differs fundamentally from methods designed for non-IID constraints. Algorithms like \textbf{FedProx}~\cite{li2020fedprox} and \textbf{SCAFFOLD}~\cite{karimireddy2020scaffold} introduce regularization terms to correct \textit{optimization drift} caused by statistical heterogeneity. Their goal is to force convergence to a global model despite divergent local updates. In contrast, DUA-D2C treats parameter divergence as a source of feature diversity. 

Furthermore, attention-based FL methods like \textbf{FedAtt}~\cite{ji2019learning} weight models based on their distance from the global model in parameter space. We argue that parameter distance is an imperfect proxy for model quality; an edge model may deviate significantly because it has learned a novel, generalizable feature, or because it has overfitted to local noise. DUA-D2C resolves this ambiguity by utilizing a shared validation set, a luxury available in a centralized setting, to weight contributions based on explicit \textit{generalization metrics} (Accuracy and Entropy) rather than geometric proximity.

\subsubsection*{Contrast with Centralized Weight Averaging}
In centralized deep learning, techniques like \textbf{Stochastic Weight Averaging (SWA)}~\cite{izmailov2018swa} aggregate weights over \textit{time} (different epochs of a single training trajectory) to find flatter minima. DUA-D2C is distinct in that it aggregates over \textit{data space} (models trained on disjoint shards). This induces a specific type of diversity driven by data isolation rather than optimization trajectory. Consequently, DUA-D2C captures a ``consensus" view that is structurally different from temporal averaging.

\subsubsection*{Key Differentiators}
We summarize the unique advantages of DUA-D2C through three key pillars:

\begin{enumerate}
    \item \textbf{Orthogonal Macroscopic Regularization:} Unlike internal regularizers such as \textbf{Dropout} or \textbf{Batch Normalization} which operate at the neuron level, DUA-D2C operates at the data level. By forcing edge models to learn from disjoint shards, we prevent the network from co-adapting to global noise patterns. Crucially, this macroscopic regularization is \textbf{orthogonal} to microscopic methods. DUA-D2C can be efficiently applied \textbf{on top of} Dropout-enabled networks, providing additive gains in generalization without conflict.
    
    \item \textbf{Quality-Aware vs. Quantity-Aware Aggregation:} Standard aggregation (e.g., \textit{FedAvg}) weights contributions based on data \textit{quantity}, assuming all data is equally valuable. DUA-D2C introduces a \textbf{quality-aware} mechanism. By leveraging the shared validation set, we compute weights based on \textit{Validation Accuracy} and \textit{Normalized Prediction Entropy}. This effectively acts as a filter that actively suppresses parameters from edge models that are ``confidently wrong" (high confidence, low accuracy) or ambiguous (high entropy), ensuring the central model only assimilates robust features.
    
    \item \textbf{Parameter Fusion Efficiency (vs. Ensembles):} Unlike \textbf{Bagging} or \textbf{Deep Ensembles}, which aggregate \textit{predictions} (voting), DUA-D2C aggregates \textit{parameters}. Prediction-based ensembles incur a high computational cost during deployment ($N \times$ inference time and memory). In contrast, DUA-D2C fuses the ``wisdom" of the subsets into a single, efficient central model. This allows for ensemble-like generalization benefits while maintaining the inference speed and deployment footprint of a single standard network.
\end{enumerate}

\section{Theoretical Framework} \label{theory}
\subsection{The Hypothesis}
Overfitting occurs when a model adapts excessively to its training data, effectively capturing and memorizing random characteristics and noise that lack significance in real-world applications. The presence of outliers and noisy data is a fundamental contributor to this phenomenon. Our approach aims to minimize the detrimental impact of such data irregularities.

The core idea is that the influence of outliers and noise can be substantially mitigated by partitioning the training data into multiple shards and training distinct models on each. In this setup, each data point is confined to a single shard. While representative samples, being closer in the feature space, are likely to be distributed somewhat uniformly across shards, individual outliers or noisy instances will primarily affect only one of the local training processes. Through periodic aggregation of these locally trained models, we can consolidate their learned knowledge.

For the basic Divide2Conquer (D2C) approach, which employs simple averaging, the adverse effect of individual outliers/noise could ideally be reduced by a factor approaching N (the number of data shards). Building upon this, our Dynamic Uncertainty-Aware D2C (DUA-D2C) introduces a more sophisticated aggregation mechanism. Instead of uniform averaging, DUA-D2C dynamically weights the contribution of each edge model based on its performance (accuracy) and predictive confidence (inverse uncertainty/entropy) on a shared validation set. This refinement aims to further amplify the benefits of the Divide2Conquer paradigm.

The key factors underpinning our hypothesis that the DUA-D2C method addresses overfitting are as follows:

\textbf{I.}
In D2C, weighted averaging of parameters combines knowledge from different subsets. Extreme parameter updates driven by noise or outliers in individual models are moderated during this averaging. If an edge model encounters a noisy sample that deviates significantly, it might adjust its parameters excessively. However, this impact is diluted when combined with parameters from other models. The DUA-D2C extension enhances this by actively down-weighting models that exhibit poor generalization (low accuracy on validation) or high uncertainty (potentially due to fitting noise in their local shard). This ensures that the aggregated parameters more reliably reflect a balanced representation of the underlying data distribution, offering a more robust mitigation against individual noisy or outlier samples.

\textbf{II.}
Averaging weights has a regularizing effect at the parameter level. D2C's averaging moderates extreme weight updates, leading to more stable parameter values. DUA-D2C refines this by making the regularization adaptive. By prioritizing models that are both accurate and confident on unseen validation data, DUA-D2C discourages the central model from adopting parameters heavily influenced by noise or idiosyncrasies from less reliable edge models. This intelligent consensus learning encourages the central model to learn more generalizable patterns and reduces reliance on individual model predictions that might be prone to overfitting their local data.

\textbf{III.}
Averaging parameters generally smooths out the decision boundaries learned by individual models. Sharp or jagged boundaries resulting from noisy updates in individual models are moderated through D2C's averaging. DUA-D2C provides an additional layer of quality control: models that likely developed overly complex or noisy decision boundaries (as indicated by poor validation performance or high uncertainty) contribute less to the central model. This leads to a central model with a potentially smoother and more robust decision boundary, better equipped for generalization to unseen data.

\textbf{IV.}
The optimal hyperparameters, such as the number of subsets $N$ and local epochs $E$, remain crucial and vary by dataset and domain. The subset count depends on dataset size and class distribution. If edge models are complex or the subset count is too high, the reduced data per subset can lead to significant overfitting in these edge models. This is similar to what happens in federated learning when the client data is minimal, as pointed out by Zhang et al. \cite{zhang2022fed}. Statistically, the variance of an estimated model $Var(\hat{f}_j)$ is often inversely related to the amount of data used for its estimation. When $n_j$ is small, the parameters $\theta_j$ learned by $\hat{f}_j$ can be highly sensitive to the specific instances in $D_j$. Small perturbations in these few samples can lead to large changes in the learned $\theta_j$, indicating high variance. The model might learn spurious correlations present only in that small $D_j$ or effectively memorize its samples, rather than capturing the underlying, generalizable patterns from the true data distribution $P(x,y)$. This local overfitting manifests as $Var(\hat{f}_j)$ being large, meaning $\hat{f}_j$ would produce widely different predictions if trained on a different small random sample from $P(x,y)$. When sufficient data exists per class ($n_j$ is reasonably large), more subsets can be beneficial, as noise or outliers are diluted by a factor of up to \textbf{N} (number of training subsets), and each $\hat{f}_j$ can learn more stable representations. While D2C's averaging helps, DUA-D2C offers a mechanism to better handle this scenario. If some edge models overfit due to limited data in their shard, their likely poor performance or high uncertainty on the validation set will reduce their influence on the central model, offering a degree of resilience against poorly performing edge models. Nevertheless, ensuring sufficient data per class in each subset remains important for all edge models to learn meaningful representations. The dynamic weighting in DUA-D2C aims to leverage the "best" of what each edge model learns, even if some are suboptimal.


In summary, DUA-D2C's dynamic uncertainty-aware aggregation helps reduce overfitting by mitigating the effects of noise and outliers and by promoting a more robust, generalized central model. While dividing the training set into too many subsets can still pose challenges, DUA-D2C provides a more sophisticated mechanism to capitalize on the contributions of potentially variably performing edge models.

\subsection{Mathematical Intuition}
To lay out the underlying intuition behind our hypothesis, we must first define overfitting formally. We follow the groundwork laid out by Ghojogh et al. \cite{ghojogh2019theory}. Let us consider a Dataset $D$ with training set $T$ and test set $R$ so that,
\begin{equation}
    T \cup R = D
    \label{eq0}
\end{equation}
\begin{equation}
    T \cap R = \emptyset
    \label{eq1}
\end{equation}
Let, $T$ consist of datapoints $\{(x_i, y_i)\}_{i = 1}^n$ and test set $R$ consist the datapoints $\{(x_i, y_i)\}_{i = 1}^m$. Also, let $f$ be the function that maps each datapoint in the training set to its corresponding output label(true observations), i.e., $f(x_i) = f_i$, where $f_i$ is the true corresponding output label for $x_i$. This model is unknown, and we wish to train a model that replicates $f$ as closely as possible. However, both the ground truth model and the true labels are unknown to us. Realistically, there is always some noise in the training dataset. We assume that the training output is corrupted with some additive noise $\epsilon_i$. So, we can say, 
\begin{equation}
    y_i = f_i + \epsilon_i 
    \label{eq2}
\end{equation}
where $\epsilon_i$ is some  Gaussian noise, $\epsilon \sim \mathcal{N}(0,\sigma^2)$. Since the mean of the distribution is zero we can say that $\mathop{{}\mathbb{E}}(\epsilon_i) = 0$, and $Var(\epsilon_i) = \sigma^2$. As the variance of a random variable $x$ can be denoted as $Var(X) = \mathop{{}\mathbb{E}}(x^2) - [\mathop{{}\mathbb{E}}(x)]^2$ (see Appendix \ref{app1}), we have,
\begin{equation}
    \mathop{{}\mathbb{E}}(\epsilon_i^2) = Var(\epsilon_i) + [\mathop{{}\mathbb{E}}(\epsilon_i)]^2  \Rightarrow \mathop{{}\mathbb{E}}(\epsilon_i^2) = \sigma^2 + 0 = \sigma^2
    \label{eq3}
\end{equation}
Our assumption is that the input of the training dataset ${\{{x}_i}\}_{i=1}^n$ and their noise-added outputs ${\{{y}_i}\}_{i=1}^n$ are available to us, and our goal is to estimate the true model $f$ by a model $\hat{f}$ in order to predict the labels ${\{{y}_i}\}_{i=1}^n$ for the input data  ${\{{x}_i}\}_{i=1}^n$. Let the estimated observations be ${\{\hat{y}_i}\}_{i=1}^n$. It is preferable that ${\{\hat{y}_i}\}_{i=1}^n$ are as close as possible to ${\{{y}_i}\}_{i=1}^n$. Mathematically, we can consider this as a minimization problem looking to minimize some error function, such as MSE (Mean Squared Error).
The MSE of estimated outputs from the trained model with respect to the dataset outputs are: 
$\textbf{MSE} = \frac{1}{n}\sum_{i=1}^n (\hat{y}_i - y_i)^2 = \mathop{{}\mathbb{E}}((\hat{y}_i - y_i)^2)$. 
Consider a single arbitrary data point $(x_0,y_0)$ and the corresponding prediction of the trained model $\hat{f}$ for the input $x_0$ is $\hat{y}_0$. Ghojogh et al. \cite{ghojogh2019theory} showed that if the sample instance $(x_0,y_0) \notin T$  i.e., it belongs to the test set $R$, MSE of the predicted label can be expressed as - 
\begin{equation}
    \textbf{MSE} =   \mathop{{}\mathbb{E}}((\hat{y}_0 - y_0)^2) = \mathop{{}\mathbb{E}}((\hat{y}_0 - f_0)^2) + \sigma^2 
    \label{eq4}
\end{equation}
Here we can observe from the first term of the R.H.S. of the equation that the MSE of the estimation of output labels with respect to the dataset labels can actually be represented by the MSE of the estimation with respect to the true uncorrupted labels ${\{{f}_i}\}_{i=1}^n$ plus some effect of the noise that exists in the dataset. Taking \textit{Monte-Carlo approximation} \cite{ghojogh2020sampling} of the expectation terms in Eq. (\ref{eq4}), we have:

\begin{align}
    \frac{1}{m} \sum_{i=1}^{m} {(\hat{y}_i - y_i)^2} = \frac{1}{m} \sum_{i=1}^{m} {(\hat{y}_i - f_i)^2} + \sigma^2 \nonumber\\ 
    \implies  \sum_{i=1}^{m} {(\hat{y}_i - y_i)^2} = \sum_{i=1}^{m} {(\hat{y}_i - f_i)^2} + m\sigma^2         
    \label{eq5}
\end{align}

Here the term $\sum_{i=1}^{m} {(\hat{y}_i - y_i)^2}$ is the total error between the model predictions and the dataset labels which we can call the \textit{empirical observed error}, $e$. On the other hand, $\sum_{i=1}^{m} {(\hat{y}_i - f_i)^2} $ represents the total error between prediction labels and the true unknown labels, namely $E$. Hence, $e = E + m\sigma^2$. Thus, the observed error truly reflects the true error because the term $m\sigma^2$ remains constant, which is the theoretical justification for evaluating model performance using unseen test data.
But, as Ghojogh et al. \cite{ghojogh2019theory} showed, if $(x_0,y_0) \in T$, deriving the M.S.E yields:
\begin{equation}
\textbf{MSE} =   \mathop{{}\mathbb{E}}((\hat{y}_0 - y_0)^2) = \mathop{{}\mathbb{E}}((\hat{y}_0 - f_0)^2) + \sigma^2 - 2\sigma^2\mathop{{}\mathbb{E}}(\frac{\partial \hat{y}_0}{\partial y_0})
\label{eq6}   
\end{equation}
Using the same approximation as Eq.(\ref{eq5}) on the training set $T$, we get:

\begin{align}
    \frac{1}{n} \sum_{i=1}^{n} {(\hat{y}_i - y_i)^2} = \frac{1}{n} \sum_{i=1}^{n} {(\hat{y}_i - f_i)^2} + \sigma^2 - 2\sigma^2 \frac{1}{n}\sum_{i=1}^{n}\frac{\partial \hat{y}_i}{\partial y_i} \nonumber\\
    \implies  \sum_{i=1}^{n} {(\hat{y}_i - y_i)^2} =  \sum_{i=1}^{n} {(\hat{y}_i - f_i)^2} + n\sigma^2 - 2\sigma^2 \sum_{i=1}^{n}\frac{\partial \hat{y}_i}{\partial y_i}
    \label{eq7}
\end{align}

\begin{figure}[ht!]
\centering
\includegraphics[width=0.6\columnwidth]{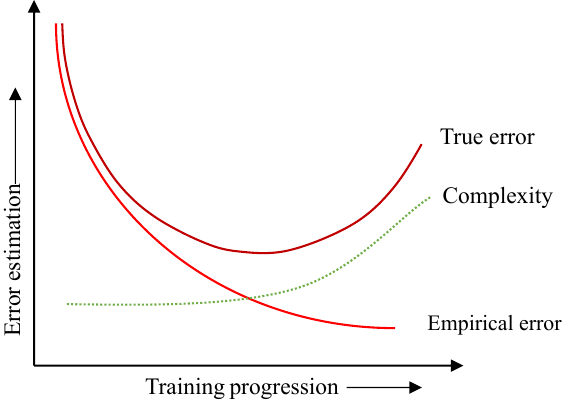}
\caption{When the corresponding complexity term grows and dominates the expression in Eq.(\ref{eq7}) for empirical error, the model starts to overfit as empirical error becomes no longer a true representation of true error.}
\label{fig: complexity-visual}
\end{figure}

\begin{figure*}[ht!]
\centering
\includegraphics[width=\columnwidth]{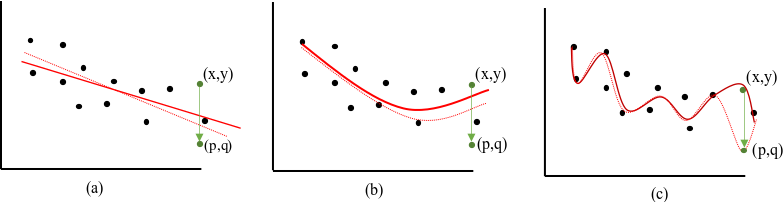}
\caption{A visualization of (a) Underfitting, (b) A good fit, and (c) Overfitting scenarios, illustrating how the model changes if an arbitrary training data point $(x,y)$ is shifted to $(p,q)$ by plotting the trained model before (solid line) and after (dashed line) the shift.}
\label{fig: overfit-visual}
\end{figure*}

From Eq.(\ref{eq7}) we now understand while the model is training the observed empirical error calculated on the training dataset i.e, training error is not a clear representation of the true error because now $e = E + n\sigma^2 - 2\sigma^2 \sum_{i=1}^{n}\frac{\partial \hat{y}_0}{\partial y_0}$. Hence, even if we can obtain a small value of $e$, the term $2\sigma^2 \sum_{i=1}^{n}\frac{\partial \hat{y}_0}{\partial y_0}$ can grow large as the training progresses and hide the true value of a substantial and large true error $E$ (Figure \ref{fig: complexity-visual}). As we can see in the figure, the model starts to overfit as empirical error becomes no longer a true representation of true error when the corresponding complexity term grows and dominates the expression in Eq.(\ref{eq7}) for empirical error. We can conclude from this analysis that the final term in Eq.(\ref{eq6}) is a measure of the overfitting of the model. Which is also known as the complexity of a model. Upon a closer look at this final term, we can derive some more interesting insight. $\frac{\partial \hat{y}_0}{\partial y_0}$ is essentially the rate of change of the predicted label $\hat{y}_i$ for the input $x_i$ with respect to the change in dataset label $y_i$. It makes sense that prediction labels would vary if the corresponding output in the dataset varies as the sample is considered from within the training set. It also reflects how dependent the model is on the training set, as the term grows when the predicted output varies too much with respect to changes in the training samples. This fits the classical definition of overfitting i.e., fitting too tightly to the training data and thus performing way worse when evaluated with actual test data. Evaluation on test data accurately represents the true error from Eq.(\ref{eq5}). Figure \ref{fig: overfit-visual} depicts a visualization of this idea.

Statistical analysis shows that the distinction between an underfitting model, an overfitting one, and a good fit can be reflected by their bias and variance with respect to the training labels. Ideally, a model with low variance and low bias is the most desirable and usually deemed a good fit. However, during the training phase, sometimes we see a model has low variance and high bias which indicates the model predictions are not good as well as being all over the place, this is the indication of an underfitting model. It usually happens when the model is too simple to capture the complexity of the data or has not been trained for long enough to reach convergence. On the other hand, we find a model that overfits has high variance and low bias. This is because the model becomes overly complex, accommodating all the outliers in the training data, and fails to generalize to unseen test data. Figure \ref{fig: overfit-visual}(a) and \ref{fig: overfit-visual}(c) show how an underfitting and overfitting model show low and high variance, respectively. Figure \ref{fig: overfit-visual} is a visualization of (a) Underfitting, (b) A good fit, and (c) Overfitting scenarios. The figures illustrate how the model changes if an arbitrary training data point $(x,y)$ is shifted to $(p,q)$ by plotting the trained model before and after the shift. In (a) the underfitted model the change is very minimal. In the (b) Good fit model the change is also relatively little, but in the (c) Overfitted model, there is a large shift in the neighboring region of that data point, which reflects the over-dependence on training data.

\subsection{Justification for the Linear Proxy}
Deep neural networks are inherently non-convex ``black boxes,'' making exact analytical derivation of their variance properties extremely difficult. To provide tractable mathematical intuition for the proposed DUA-D2C framework, we have adopted a simplified linear regression model as an analogue in the following derivations. This is a standard practice in learning theory; seminal works on ensemble methods (e.g., Bagging \cite{breiman1996bagging}) often utilize linear estimators to illustrate the mechanics of bias-variance decomposition. While deep networks are non-linear globally, they can be well-approximated by linear models locally in the parameter space, particularly during the fine-tuning stages or under the Neural Tangent Kernel (NTK) regime \cite{jacot2018neural}. Therefore, the variance reduction principles derived from this linear proxy provide a valid first-order approximation for the behavior of deep networks under aggregation.

\subsection{Analysis of Variance Reduction in D2C}
To see the impact of \textbf{Divide2Conquer} we consider the training set $T$ divided into $k$ equal partitions, namely $\lvert T_1\rvert = \lvert T_2\rvert=...=\lvert T_k\rvert = \lvert T \rvert/k$. Then, we train the model $\hat{f}_j$ using the $j$-th data chunk, where $j \in \{1,..,k\}$. As a result, we have $k$ separately trained models. Finally, we aggregate these edge models into one central model $\hat{f}$. This process is repeated for $p$ global epochs. For the sake of simplicity, we consider each $\hat{f}_j$ is a regression model with weight matrix $\pmb{W_j} = \{w_{j1},...,w_{jn}\}$ and bias $b_j$. For an input vector  $\pmb{x} = \{x_1,...,x_n\}$ the prediction of the edge model on the $t$-th epoch can be described as: $\hat{f}_j(\pmb{x}) = \pmb{W_j}^t\pmb{x}^T+b_j$. In the case of a single global epoch, we now wish to aggregate the weight matrices creating the central weight matrix $\pmb{\Bar{W}} = \{\Bar{w}_{1},...,\Bar{w}_{n}\}$. Here, the weight matrix for the $(t+1)$-th epoch is: 
 \begin{equation}
 \begin{aligned}
   \pmb{\Bar{W}}^{t+1}  = \frac{1}{k} \sum_{j=1}^{k} \pmb{W_j}^{t}  \textrm{  and  } b^{t+1} = \frac{1}{k} \sum_{j=1}^{k} {b^t_{j}}
    \label{eq8} 
 \end{aligned}
 \end{equation}
 Hence, the output of the central model at $(t+1)$-th epoch:

 \begin{align}
      &\hat{f}(\pmb{x})^{t+1}  = \pmb{\Bar{W}}^{t+1}\pmb{x}^T+b^{t+1} \nonumber\\
      &\implies \hat{f}(\pmb{x})^{t+1}  =  \frac{1}{k} \sum_{j=1}^{k} \pmb{W_j}^{t}\pmb{x}^T + \frac{1}{k} \sum_{j=1}^{k} {b^t_{j}} \nonumber \\
      &\implies \hat{f}(\pmb{x})^{t+1}  =  \frac{1}{k}   \sum_{j=1}^{k} \left( \pmb{W_j}^{t}\pmb{x}^T +  {b^t_{j}} \right) \nonumber \\
      &\implies \hat{f}(\pmb{x})^{t+1} = \frac{1}{k}   \sum_{j=1}^{k} \hat{f}(\pmb{x})^{t}_j 
     \label{eq9}
  \end{align}

Let $e_{jt}$ denote the empirical error of the $j$-th model while estimating some arbitrary instance on the $t$-th global epoch. We assume this error to be a random variable with normal distribution having mean zero, i.e., $e_{jt} \sim \mathcal{N}(0,s)$. Hence, $\mathop{{}\mathbb{E}}(e_{jt}) = 0 $ and $Var(e_{jt}) = s $. Let us denote the estimation of the product of two trained models using two different data chunks by $c$, i.e., $ \mathop{{}\mathbb{E}}(e_{jt} \times e_{lt}) = c$. So, we have:
\begin{equation}
 \begin{aligned}
    \mathop{{}\mathbb{E}}(e_{jt}^2) = Var(e_{jt}) + [\mathop{{}\mathbb{E}}(e_{jt})]^2  \Rightarrow \mathop{{}\mathbb{E}}(e_{jt}^2) = s + 0 = s
    \label{eq10} 
 \end{aligned}
\end{equation}

Again for $j,l \in \{1,...,k\}$, where $j\neq l$, we have:
\begin{align}
     &Cov(e_{jt} ,e_{lt}) = \mathop{{}\mathbb{E}}(e_{jt} \times e_{lt}) - \mathop{{}\mathbb{E}}(e_{jt})\mathop{{}\mathbb{E}}(e_{lt}) \nonumber\\
     &\implies Cov(e_{jt} ,e_{lt}) = c - 0\times0 = c \nonumber\\
     &\implies Cov(\hat{f}(\pmb{x})^{t}_j ,\hat{f}(\pmb{x})^{t}_l) = c - 0\times0 = c
    \label{eq11} 
\end{align}
 
From Eq.(\ref{eq9}) and Eq.(\ref{eq11}), we can deduce:

\begin{equation}
  \begin{aligned}
     Var(\hat{f}(\pmb{x})^{t+1}) &= \frac{1}{k^2} Var\left(\sum_{j=1}^{k} \hat{f}(\pmb{x})^{t}_j\right)\\ 
     &=\frac{1}{k^2} \sum_{j=1}^{k} Var\left(\hat{f}(\pmb{x})^{t}_j\right) + \frac{1}{k^2} \sum_{i=1,l=1}^{k} \\ &\sum_{i\neq l}^{k}Cov\left(\hat{f}(\pmb{x})^{t}_i, \hat{f}(\pmb{x})^{t}_l\right) \\ 
     &= \frac{1}{k^2} sk + \frac{1}{k^2} ck(k-1) =  \frac{s+c(k-1)}{k} 
         \label{eq12} 
  \end{aligned}
\end{equation}

Eq.(\ref{eq12}) provides us with an interesting insight. If the two edge models are highly correlated, i.e., their predictions and empirical error are very close to each other, we can assume $s\approx c$. Hence, the variance of the central model at $(t+1)$-th epoch can be reduced to: $Var(\hat{f}(\pmb{x})^{t+1}) \approx  (s+s(k-1))/k \approx s $. On the other hand, if the two models are very dissimilar, the same expression gives us:  $Var(\hat{f}(\pmb{x})^{t+1}) \approx  (s+0(k-1))/k \approx s/k $. As $Var(e_{jt}) = Var(\hat{f}(\pmb{x})^{t})= s$, each of the edge models at epoch $t$ has a variance of prediction $s$, which means if the edge models at epoch $t$ are completely different, the variation of estimate decreases by a factor of $k$ on epoch $(t+1)$. Consequently, if the edge models are identically trained, the variance remains almost unchanged. This result is similar to the one proposed by Breiman et al. \cite{breiman1996bagging} and B{\"u}hlmann et al. \cite{buhlmann2002analyzing} through the idea of \textit{'Bagging'}, which is a meta-algorithm aggregating multiple model outputs. The analysis clearly shows that our approach does not at least worsen the problem of overfitting in this scenario, but rather improves the central model on each epoch if certain conditions are met. However, in reality, the edge models are neither completely different nor exactly the same due to the relative distribution of the data subsets and their shared model architecture. So, the degree to which D2C ameliorates overfitting depends on the model architecture and the correlation among the edge models. However, it is easy to see that if the training data is contaminated by outliers and random noise, the learned parameters from the different portions of the data are more likely to be different from each other. This is because the individual outliers and noise are not highly correlated, distributing them among the subsets likely results in significantly different samples in the subsets, making the subsequent models different from each other. 

\subsection{Extension to Non-Linear Deep Networks}
While the previous derivation assumes linearity, we formally extend this intuition to non-linear deep networks via \textbf{local linearization} in the parameter space. This approach relies on the theoretical observation that wide neural networks, particularly in the Neural Tangent Kernel (NTK) regime \cite{jacot2018neural}, exhibit training dynamics where parameters remain close to their initialization, allowing the network to be well-approximated by a linearized model.

Let $\hat{f}(x; \theta)$ be a non-linear deep neural network parameterized by $\theta \in \mathbb{R}^d$. We assume the trained parameters of the $j$-th edge model, $\theta_j$, remain within a local neighborhood of a reference configuration $\theta^*$ (e.g., the central model initialization). We can approximate the output of the $j$-th edge model using a first-order Taylor expansion around $\theta^*$:

\begin{equation}
    \hat{f}(x; \theta_j) \approx \hat{f}(x; \theta^*) + \nabla_\theta \hat{f}(x; \theta^*)^\top (\theta_j - \theta^*)
    \label{eq:taylor_individual}
\end{equation}

In D2C, the central model parameters $\theta_{central}$ are obtained by averaging the edge parameters:
\begin{equation}
    \theta_{central} = \frac{1}{k} \sum_{j=1}^k \theta_j
\end{equation}

To determine the output of the central model, we apply the same linear approximation to $\hat{f}(x; \theta_{central})$:

\begin{equation}
    \hat{f}(x; \theta_{central}) \approx \hat{f}(x; \theta^*) + \nabla_\theta \hat{f}(x; \theta^*)^\top (\theta_{central} - \theta^*)
\end{equation}

Substituting $\theta_{central} = \frac{1}{k} \sum_{j=1}^k \theta_j$:

\begin{equation}
\begin{aligned}
    \hat{f}(x; \theta_{central}) &\approx \hat{f}(x; \theta^*) + \nabla_\theta \hat{f}(x; \theta^*)^\top \left( \frac{1}{k} \sum_{j=1}^k \theta_j - \theta^* \right) \\
    &= \hat{f}(x; \theta^*) + \frac{1}{k} \sum_{j=1}^k \nabla_\theta \hat{f}(x; \theta^*)^\top (\theta_j - \theta^*)
\end{aligned}
\end{equation}

Since $\hat{f}(x; \theta^*)$ is a constant with respect to the summation, we can distribute it into the sum (noting that $\hat{f}(x; \theta^*) = \frac{1}{k} \sum_{j=1}^k \hat{f}(x; \theta^*)$):

\begin{equation}
\begin{aligned}
    \hat{f}(x; \theta_{central}) &\approx \frac{1}{k} \sum_{j=1}^k \left[ \hat{f}(x; \theta^*) + \nabla_\theta \hat{f}(x; \theta^*)^\top (\theta_j - \theta^*) \right] \\
    &\approx \frac{1}{k} \sum_{j=1}^k \hat{f}(x; \theta_j) \quad \text{(using Eq. \ref{eq:taylor_individual})}
\end{aligned}
\label{eq:equivalence}
\end{equation}

Eq. (\ref{eq:equivalence}) proves that under local linearization, \textbf{averaging the parameters is mathematically equivalent to averaging the function outputs}. Consequently, the variance reduction properties derived in Eq. (\ref{eq12}) for the linear case apply directly to the predictions of the central deep network. By averaging the weights of $k$ edge models trained on different data shards, we effectively construct an ensemble prediction that reduces the variance of the estimator, thereby smoothing the decision boundary and mitigating overfitting even in non-linear architectures.

\subsection{Weighted Aggregation in DUA-D2C}
DUA-D2C refines this by introducing dynamic weights $\alpha_j$ based on validation accuracy and inverse entropy. The aggregated parameters are $\pmb{\Bar{W}}^{t+1} = \sum_{j=1}^{k} \alpha_j \pmb{W_j}^{t}$. 
In a realistic scenario, edge models exhibit heterogeneous variance $\sigma_j^2$ (rather than a uniform $s$) depending on the noise level of their specific data shard. The variance of the weighted sum is:
\begin{equation}
 \begin{aligned}
    Var(\hat{f}_{DUA}) &= \sum_{j=1}^{k} \alpha_j^2 Var(\hat{f}_j) + \sum_{i \neq l} \alpha_i \alpha_l Cov(\hat{f}_i, \hat{f}_l) \\
    &= \sum_{j=1}^{k} \alpha_j^2 \sigma_j^2 + \sum_{i \neq l} \alpha_i \alpha_l c_{il}
        \label{eq_dua_variance} 
 \end{aligned}
\end{equation}
Statistical theory dictates that to minimize the variance of a weighted sum of estimators with different variances, the weights $\alpha_j$ should be inversely proportional to the individual variances ($\alpha_j \propto 1/\sigma_j^2$).

In DUA-D2C, the composite score serves as an empirical proxy for this inverse variance. A model that has overfitted to noise will exhibit high prediction entropy or poor validation accuracy, signaling a high intrinsic variance $\sigma_j^2$ with respect to the true distribution. DUA-D2C assigns a low weight $\alpha_j$ to such models. 
Mathematically, if an edge model $k$ has high variance ($\sigma_k^2 \gg \sigma_{others}^2$), simple averaging (where $\alpha_k = 1/N$) would allow the term $\frac{1}{N^2}\sigma_k^2$ to significantly inflate the total variance. DUA-D2C, by driving $\alpha_k \to 0$, effectively removes this high-variance term from the summation. This ensures that the central model converges closer to the true underlying function $f_0$ by filtering out the contribution of noisy shards.

In summary, the D2C framework leverages the diversity induced by data partitioning to reduce the variance of the aggregated model, effectively diluting the impact of localized noise. The DUA-D2C extension advances this paradigm by implementing a dynamic, uncertainty-aware weighting scheme that serves as an empirical proxy for inverse variance weighting. By preferentially integrating parameters from reliable, well-generalizing edge models, DUA-D2C minimizes the effective variance injected into the central model, resulting in superior robustness against overfitting. This analysis provides a rigorous mathematical foundation for the ``divide and conquer'' strategy, confirming that adaptive aggregation is a sound mechanism for enhancing generalization in complex deep learning architectures.

\section{Proposed Methodology} \label{method}

This study proposes the \textbf{Dynamic Uncertainty-Aware Divide2Conquer (DUA-D2C)} framework. While building upon the structural foundation of data partitioning, DUA-D2C introduces specific methodological innovations that distinguish it from standard ensemble or federated approaches. Specifically, we propose:

\begin{enumerate}
    \item \textbf{Dynamic Quality Assessment:} Unlike static averaging, we introduce an intermediate evaluation phase where every edge model is assessed on a shared validation set for both generalization accuracy and predictive entropy.
    \item \textbf{Uncertainty-Aware Filtering:} We formulate a novel weighting mechanism (Eq. \ref{eq:composite_score}) that uses \textit{normalized inverse entropy} to actively suppress the contribution of models that are ``confidently wrong" or highly uncertain, effectively filtering out noise induced by data partitioning.
    \item \textbf{Parameter-Space Fusion:} We aggregate models in the parameter space rather than the prediction space, allowing for ensemble-like robustness without the inference latency of traditional bagging.
\end{enumerate}

To implement these innovations, the methodology is structured into four distinct phases: data partitioning, local edge model training, dynamic evaluation and weighting, and uncertainty-aware global aggregation.

\subsection{Phase 1: Creating Training Subsets} \label{subsec:subsets}
The initial phase prepares the data for distributed processing. The training dataset $D$ is randomly shuffled to ensure an unbiased distribution and then partitioned into $N$ distinct, non-overlapping subsets (shards): $\{D_1, D_2, \dots, D_N\}$. A critical consideration is maintaining class balance; we ensure that class proportions within each subset $D_i$ mimic the overall dataset $D$. Each subset is assigned to a dedicated \textbf{edge model}, $M_i$. All $N$ edge models and the central model $M_c$ share an identical architecture and are initialized with the same parameters to facilitate coherent aggregation.

\subsection{Phase 2: Local Training of Edge Models} \label{subsec:local_training}
This phase constitutes the inner loop of the DUA-D2C process. Each edge model $M_i$ is trained exclusively on its assigned subset $D_i$ for a fixed number of local epochs ($E$). Upon completion of $E$ epochs, each edge model $M_i$ possesses an updated set of parameters, $\theta'_i$, which reflects the features learned from its specific data shard.

\subsection{Phase 3: Edge Model Evaluation and Dynamic Weighting} \label{subsec:evaluation_weighting}
This phase contains the core novelty of DUA-D2C. Unlike static averaging, we evaluate each edge model $M_i$ (with parameters $\theta'_i$) on a shared validation dataset, $D_{val}$, to compute two metrics:
\begin{enumerate}
    \item \textbf{Generalization:} Validation accuracy, $a_i$.
    \item \textbf{Uncertainty:} Average prediction entropy, $\epsilon_i$, over $D_{val}$.
\end{enumerate}
To construct a usable weighting metric, we calculate the raw inverse entropy $inv\_e_i = 1 / (\epsilon_i + \delta_e)$ and cap it at $C_{max}$ to prevent outliers from dominating the aggregation ($inv\_e_{cap_i} = \min(inv\_e_i, C_{max})$). This capped value is Min-Max scaled to $[0,1]$ to yield a \textbf{normalized confidence score}, $u_i$:
\begin{equation}
    u_i = \frac{inv\_e_{cap_i} - inv\_e_{min}}{C_{max} - inv\_e_{min} + \epsilon_s}
    \label{eq:confidence_score}
\end{equation}
where $inv\_e_{min}$ is the theoretical minimum inverse entropy. Finally, we compute a composite raw scaling factor $s_i$:
\begin{equation}
    s_i = \lambda \cdot a_i + (1 - \lambda) \cdot u_i
    \label{eq:composite_score}
\end{equation}
Here, $\lambda$ is a hyperparameter controlling the trade-off between accuracy and uncertainty.

\subsection{Phase 4: Uncertainty-Aware Global Aggregation} \label{subsec:aggregation}
In the outer loop, the raw scaling factors are normalized to produce aggregation weights $\alpha_i = s_i / (\sum s_j + \epsilon_s)$. The central model parameters $\theta_c$ are updated via weighted averaging:
\begin{equation}
    \theta_c \leftarrow \sum_{i=1}^{N} \alpha_i \cdot \theta'_i
    \label{eq:weighted_aggregation}
\end{equation}
This ensures that models with superior generalization and lower uncertainty contribute more to the central model. The updated $\theta_c$ is then redistributed to all edge models for the next global epoch. The complete procedure is detailed in Algorithm \ref{alg:dua-d2c}.

\begin{algorithm}[!htbp]
\caption{Dynamic Uncertainty-Aware Divide2Conquer (DUA-D2C)}\label{alg:dua-d2c}
\begin{algorithmic}[1]
\STATE \textbf{Input:} Training data $D$, Validation data $D_{val}$; Num. subsets $N$; Epochs $E, E_{global}$; Batch $B$; Learning rate $lr$; Parameters: $\lambda, C_{max}, \delta_e, K, \epsilon_s$
\STATE \textbf{Output:} Central model $M_c$ with aggregated weights $\theta_c$
\STATE
\STATE \textbf{function} DUA-D2C($D, D_{val}, N, E, E_{global}, B, lr, \lambda, C_{max}, \delta_e, K, \epsilon_s$)
\STATE \hspace{0.5cm} Shuffle $D$; Divide $D$ into $N$ subsets: $\{D_1, \dots, D_N\}$ maintaining class distributions
\STATE \hspace{0.5cm} Initialize central model $M_c$ with weights $\theta_c$
\STATE \hspace{0.5cm} Set min. inverse entropy: $inv\_e_{min} = 1 / (\log(K) + \delta_e)$
\FOR{$g = 1$ \textbf{to} $E_{global}$}
    \STATE \hspace{0.5cm} $W_{central} = 0$; $S_{raw} = [\,]$
    \STATE \hspace{0.5cm} Let $\Theta'_{edge} = [\,]$ be a list to store edge model weights
    \FOR{$i = 1$ \textbf{to} $N$}
        \STATE \hspace{1cm} Initialize edge model $M_i$ with $\theta_i = \theta_c$
        \FOR{$e = 1$ \textbf{to} $E$}
            \STATE \hspace{1.5cm} Train $M_i$ on $D_i$ (batch $B$, learning rate $lr$)
        \ENDFOR
        \STATE
        \STATE \hspace{1cm} Compute $a_i$ (accuracy of $M_i$ on $D_{val}$)
        \STATE \hspace{1cm} Compute $\epsilon_i$ (prediction entropy of $M_i$ on $D_{val}$)
        \STATE \hspace{1cm} $inv\_e_i = 1 / (\epsilon_i + \delta_e)$
        \STATE \hspace{1cm} $inv\_e_{cap_i} = \min(inv\_e_i, C_{max})$
        \STATE \hspace{1cm} $u_i = (inv\_e_{cap_i} - inv\_e_{min}) / (C_{max} - inv\_e_{min} + \epsilon_s)$
        \STATE \hspace{1cm} $u_i = \max(0, \min(1, u_i))$ \COMMENT{Clamp to $[0,1]$}
        \STATE \hspace{1cm} $s_i = \lambda \cdot a_i + (1 - \lambda) \cdot u_i$
        \STATE \hspace{1cm} Append $s_i$ to $S_{raw}$
        \STATE \hspace{1cm} Obtain updated weights $\theta'_i$ from $M_i$; Append $\theta'_i$ to $\Theta'_{edge}$
    \ENDFOR
    \STATE
    \STATE \hspace{0.5cm} $Sum\_S_{raw} = \sum_{j=1}^{N} S_{raw}[j]$
    \STATE \hspace{0.5cm} Let $S_{norm} = [\,]$ be a list for normalized scaling factors
    \FOR{$i = 1$ \textbf{to} $N$}
        \STATE \hspace{1cm} $S_{norm}[i] = S_{raw}[i] / (Sum\_S_{raw} + \epsilon_s)$
    \ENDFOR
    \STATE
    \FOR{$i = 1$ \textbf{to} $N$}
        \STATE \hspace{1cm} $\theta_{scaled_i} = S_{norm}[i] \cdot \Theta'_{edge}[i]$
        \STATE \hspace{1cm} $W_{central} = W_{central} + \theta_{scaled_i}$
    \ENDFOR
    \STATE
    \STATE \hspace{0.5cm} $\theta_c = W_{central}$ \COMMENT{Update central model weights}
    \STATE \hspace{0.5cm} Update $M_c$ with $\theta_c$
\ENDFOR
\STATE
\STATE \textbf{return} central model $M_c$
\STATE \textbf{end function}
\end{algorithmic}
\end{algorithm}

\subsection{Sensitivity Analysis and Parameter Selection} \label{subsec:hyper_tuning}
To ensure that DUA-D2C achieves the optimal generalization, a sensitivity analysis should be performed on the newly introduced global hyperparameters. Improper parameter selection can introduce specific uncertainties into the aggregation process:

\begin{itemize}
    \item \textbf{Number of Subsets ($N$):} This is the most influential parameter. Increasing $N$ reduces the theoretical variance of the central model (as per Eq. \ref{eq12}) but simultaneously reduces the data size ($|D|/N$) available for each edge model. If $N$ is too large, edge models may suffer from high variance due to data scarcity, introducing noise into the aggregation. We tune $N$ by observing validation loss; typically, an optimal $N$ exists where the benefit of diversity outweighs the cost of data reduction.
    \item \textbf{Number of Local Epochs ($E$):} The number of epochs each edge model trains on its subset before aggregation. A small number (e.g., $1 \sim 3$) is recommended, as edge models can quickly overfit their reduced-size training subset.
    \item \textbf{Tradeoff Parameter ($\lambda$):} This parameter ($\in [0,1]$) controls the influence of validation accuracy ($a_i$) versus confidence ($u_i$). If $\lambda \to 1$, the model ignores uncertainty, potentially favoring ``lucky" models that are accurate but unconfident. If $\lambda \to 0$, the model favors pure confidence, risking the amplification of ``confidently wrong" predictions. Sensitivity analysis suggests a balanced range ($0.5 \sim 0.8$) provides the most robust noise filtering.
    \item \textbf{Max Inverse Entropy Cap ($C_{max}$):} Without this cap, a single edge model with near-zero entropy (infinite inverse entropy) could dominate the weighted average, effectively reverting the ensemble to a single model. $C_{max}$ introduces a ceiling to ensure diversity is preserved in the aggregation.
\end{itemize}

The tuning process involves a hierarchical search: identifying a stable range for $N$ and $E$ using the baseline D2C first, and then fine-tuning $\lambda$ and $C_{max}$ for DUA-D2C to maximize generalization on the validation set.

\section{Experimental Setup}  \label{exp}
\subsection{Datasets}
We selected a range of benchmark datasets from three different domains to test and validate our proposed method. For image classification tasks, we employed three widely used benchmark datasets: CIFAR-10 \cite{krizhevsky2014cifar}, MNIST \cite{lecun1998gradient}, and Fashion MNIST \cite{xiao2017fashion}. CIFAR-10 provides a challenging image classification task with 60,000 $32\times32$ RGB images across 10 categories. MNIST offers a simpler task with 70,000 $28\times28$ grayscale images of handwritten digits. Fashion MNIST presents a more complex variation with 70,000 $28\times28$ grayscale images of fashion items. To address class imbalance and the overfitting challenge in facial expression recognition, we utilized the FER-2013 \cite{zahara2020facial} dataset, which comprises over 35,000 grayscale images of seven basic emotions. The diverse and noisy nature of this internet-sourced dataset makes it particularly suitable for evaluating generalization performance. For audio-based tasks, we created a comprehensive dataset by combining TESS \cite{dupuis2010toronto}, CREMA-D \cite{cao2014crema}, and RAVDESS \cite{livingstone2018ryerson}. This dataset encompasses 14 emotion classes, offering a diverse range of emotional expressions and recording conditions. By merging these datasets, we aim to improve model generalization by mitigating biases and limitations inherent in individual datasets. To test our method on textual data, we employed the AG News \cite{zhang2015character} dataset, which consists of approximately 127,600 news articles categorized into four balanced classes: World, Sports, Business, and Science/Technology. This dataset provides a large-scale, real-world benchmark for evaluating text classification models, particularly in the context of Big Data.

\subsection{Experimental Details}
\textbf{1. Image Classification:} At first, all the images are reshaped and normalized. We applied one-hot encoding to the labels to use cross-entropy loss later. We split the training data into training and validation sets using Scikit-Learn. We then divided the training set into multiple subsets. Before constructing the model, processing each training subset into tensors, and batching them was our last step. Each of our edge models was comprised of 4 Convolutional (32 to 256 kernels progressively) and MaxPooling layer (2X2) blocks, followed by one fully connected hidden layer (128 neurons). Each block contained two identical convolutional layers and one pooling layer.

\textbf{2. Audio Classification: }At first, we combined the paths of the files of our three datasets into a single CSV file. Then we read the WAV files as time series and applied white noise. Then we converted the 1D time series into 2D Log Mel Spectrograms to capture the audio signals' time and frequency domain features. We treated the resulting data as 2D images, and then the remaining steps were the same as we mentioned in the case of the Image classification tasks.

\textbf{3. Text Classification:} In this case, we created a corpus using all the words occurring in our training dataset. Then we made a word index and word embedding for all the words in the corpus. After that, we created word sequences using that word embedding for each data instance. We applied padding to make each training data point equal in length. In the case of these 1D text sequences, the edge model we used consisted of 3 bidirectional LSTM layers(128, 64, and 32 units) and two dense layers(128 and 64 neurons) followed by the output layer. It took the embeddings as input, and we used the 100-dimensional version of GloVe from Stanford for these embeddings.

In all cases, we used a 90-10 Training-Validation split, Adam optimizer, and applied dropouts and batch normalization layers. This is important to test for DUA-D2C's effectiveness when used on top of other methods that address overfitting. We used the predefined test sets for testing in all cases, and nested validation sets are also recommended as best practise. 

\section{Results and Discussions}  \label{result}
We will use loss and accuracy curves, Accuracy, F1 Score, MCC, Cohen's Cappa, AUC-ROC, and log loss to analyze the results and evaluate our method. We also varied the two new global hyperparameters we mentioned in the previous section: the \textbf{Number of Subsets of the Training Set} and the \textbf{Number of Epochs before each round of Central Averaging}. To refer to them concisely, we will use the variables $\pmb{N}$ and $\pmb{E}$ respectively in this section. As for the Tradeoff Parameter ($\lambda$), we used a value of 0.7 in each case for the sake of consistency, and due to the fact that we found this value to be a good one.

\subsection{Visualizing Decision Boundary}
At first, we will visualize the change in decision boundaries due to applying our proposed method and analyze the findings. We used a synthetic binary classification dataset to simulate a simple decision boundary. This dataset was created using the Scikit-Learn library. Then the 240-sample large dataset was divided into a training and a test set using a 75-25 split. The neural network architecture we used for these experiments comprised three hidden layers having 100 neurons each. No regularization or data augmentation was used for decision boundary visualisation, to observe the effect of the Divide2Conquer approach explicitly. For decision boundary analysis, we kept things simple and used the basic D2C approach without performance-based aggregation.

At first, we adopted the centralized, traditional approach and fed the training dataset to the sole model. After training, we got the decision boundary shown in Figure \ref{fig: n1-dbl}.

\begin{figure*}[t]
  \centering
  \begin{subfigure}[t]{0.49\textwidth}
    \centering
    \includegraphics[width=\linewidth]{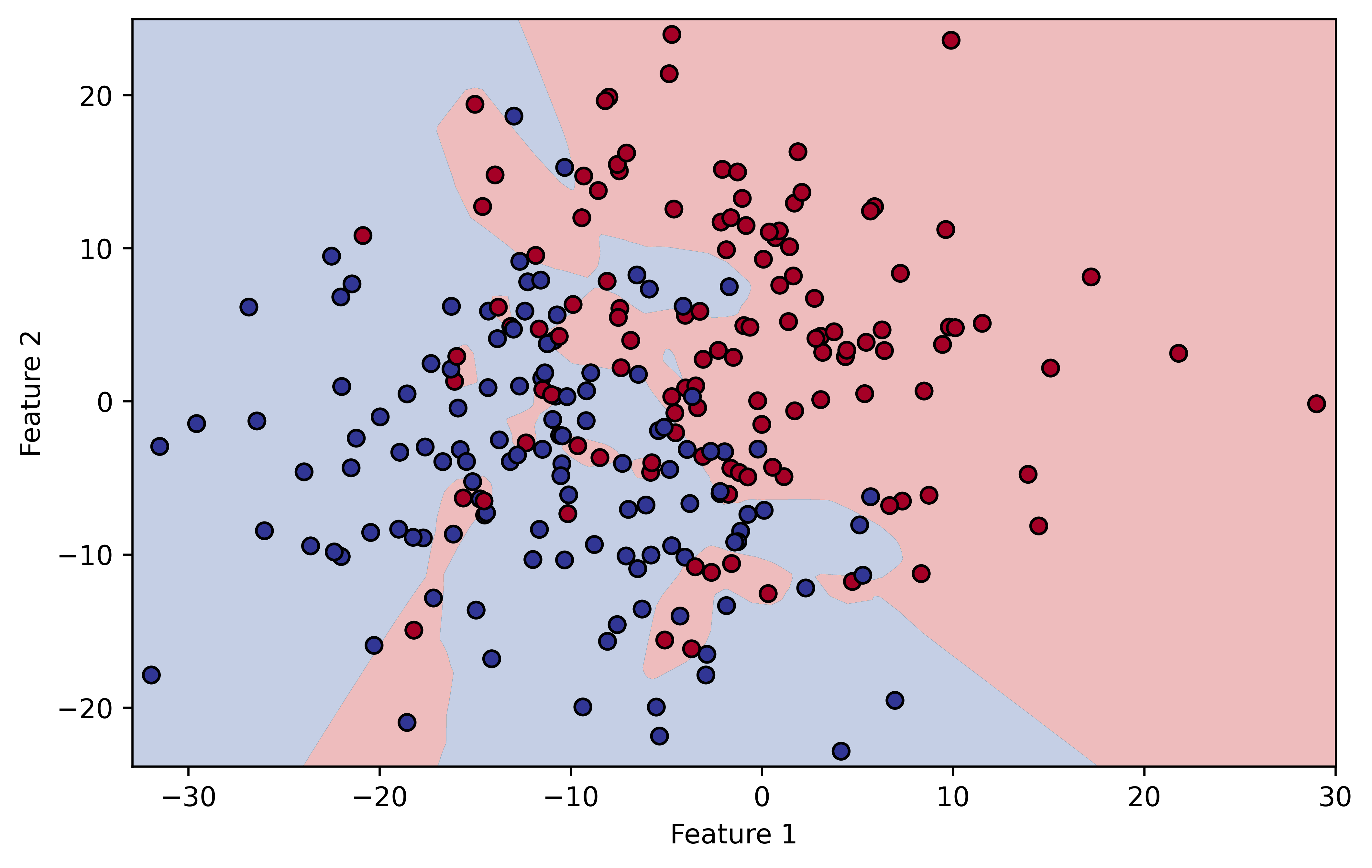}
    \caption{No Subsets (Traditional)}
    \label{fig:n1-dbl}
  \end{subfigure}
  \hfill
  \begin{subfigure}[t]{0.478\textwidth}
    \centering
    \includegraphics[width=\linewidth]{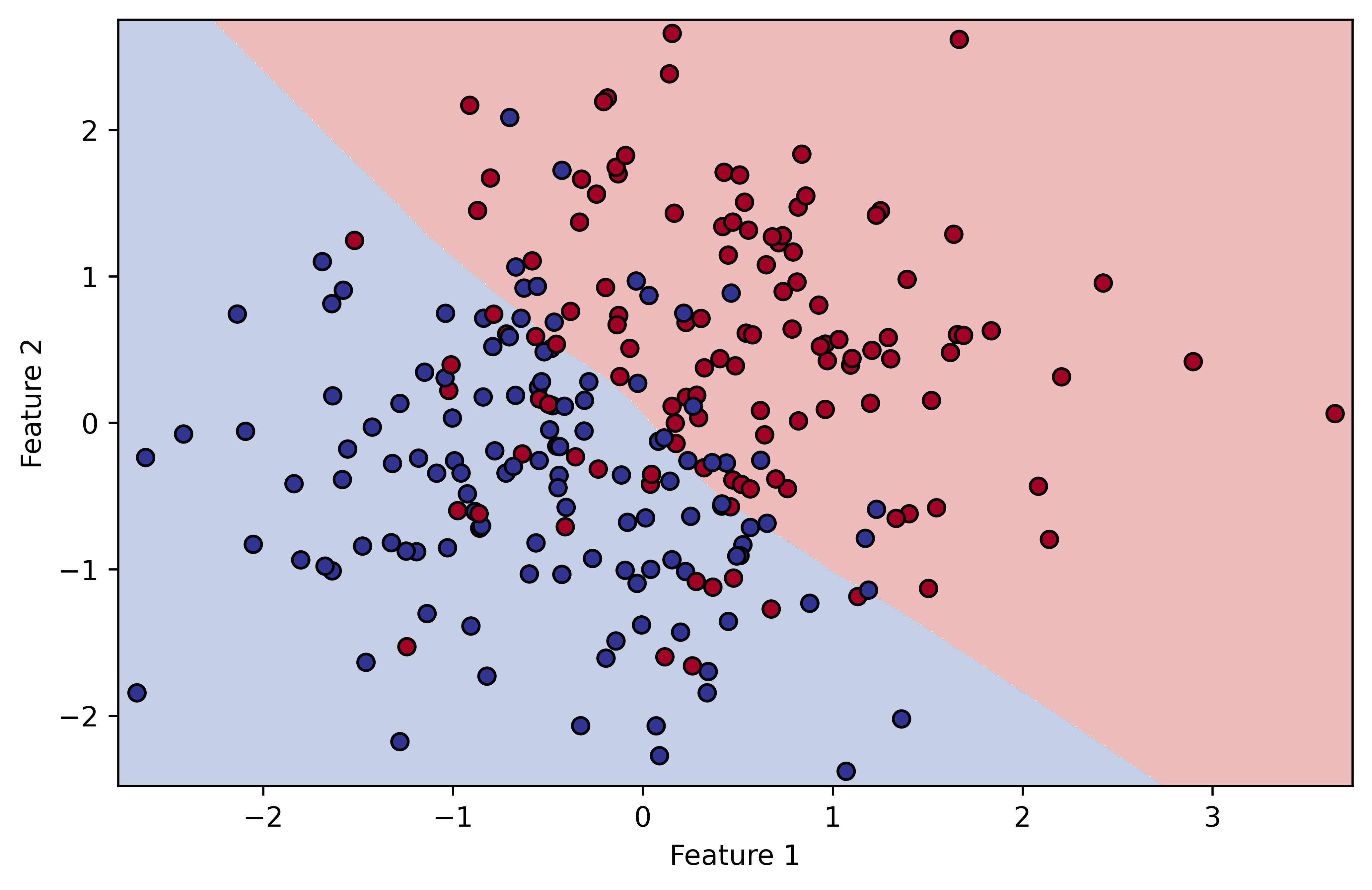}
    \caption{N = 3 (D2C)}
    \label{fig:n3-db}
  \end{subfigure}
  \caption{Decision boundary visualization with the Traditional approach vs. Divide2Conquer (D2C).}
  \label{fig:decision-boundary}
\end{figure*}

\begin{figure*}[ht!]
\centering
\includegraphics[width=1\linewidth]{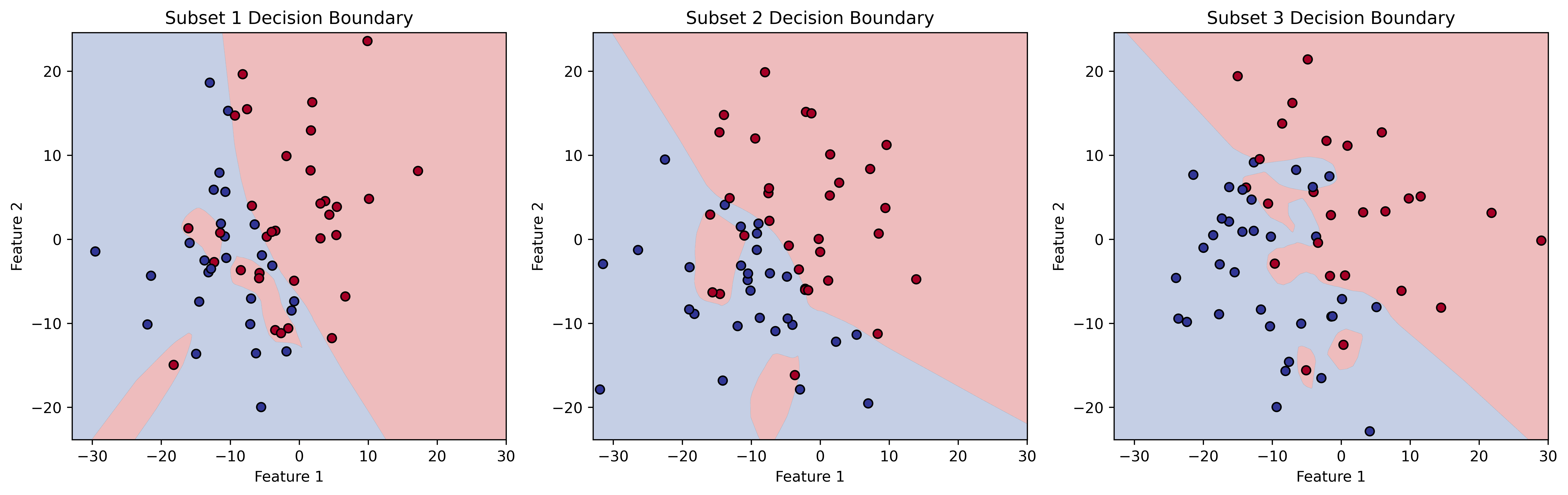}
\caption{Decision Boundary from the edge models trained on each subset after dividing the training data into 3 subsets using the proposed method.}
\label{fig: subset-dbl}
\end{figure*}

As we can see, the decision boundary for the traditional approach is highly complex and non-linear here. The model creates intricate regions to distinguish between the two classes. The boundary closely follows individual points, resulting in a convoluted, jagged shape. The complexity of the decision boundary suggests that the model is overfitting to the training data. It is trying too hard to separate every point, including noise, rather than focusing on general patterns in the data. Evidence of overfitting is seen in how the model creates small pockets of blue in the red region and vice versa. This indicates the model has likely memorized the data rather than learning generalizable patterns. This means that this approach is likely to produce excellent results on training data. Still, it is likely to struggle with new, unseen data due to its sensitivity to specific details in the training set. The highly irregular boundary reflects poor generalization ability. 

Then, to see the effect of D2C on decision boundaries, we divided our training set into multiple subsets and trained them in parallel using identical \textbf{Edge Models}. Figure \ref{fig: subset-dbl} shows the decision boundaries we got from each edge model. 

As we can see, the decision boundaries from the edge models are still complex, containing intricate regions following individual noisy data points. However, each of these noisy samples occurs only once in one of the subsets, affecting the corresponding edge model. So, irregularity around an individual noisy sample does not affect all the edge models but rather only one of them. Hence, after averaging the weights, the impact of the individual outliers should be diluted. 

To verify, the trained weights from each subset were averaged to get the final weights for the \textbf{Central Model}. The resulting decision boundary is shown in Figure \ref{fig: n3-db}.

We can see that the decision boundary from the central model of our method is much simpler, dividing the feature space almost diagonally. This suggests a less flexible, more generalized model due to averaging the weights from multiple subsets. This boundary is smooth and does not follow the exact positions of individual data points as closely. The model is more generalized, focusing on the overall trend in the data rather than trying to accommodate every single datapoint, exhibiting much less overfitting compared to what we saw in the traditional approach using the same model architecture. By averaging the weights from different subsets, the model smooths out the noise in the data and prevents overfitting to specific training samples. An individual noisy sample does not affect all the edge models in D2C, and its impact is diluted through averaging. This leads the central model to create a decision boundary based on the more common patterns present in all the subsets, ensuring better generalization. Further proof of this generalization is seen in the training and test accuracies for the traditional approach and D2C using different numbers of subsets. We summarize these accuracies in Table \ref{tab:db-tab}. 

\begin{table}[ht]
\renewcommand{\arraystretch}{1.3}
\centering
\caption{\label{tab:db-tab}Performance of the traditional approach vs different values of $N$ Using the proposed method.}
\begin{tabular}{|c|c|c|}
\hline
\textit{Model Settings}                    & \textit{Training Accuracy} & \textit{Test Accuracy}   \\ \hline \hline
No Subsets & \textbf{0.9944}   & 0.6833          \\ 
N=2 (2 Subsets)                   & 0.7611            & 0.75            \\ 
N=3 (3 Subsets)                   & 0.7944            & \textbf{0.8333} \\ 
N=4 (4 Subsets)                   & 0.7889            & \textbf{0.8333} \\ \hline
\end{tabular}
\end{table}

As we can see, we got almost perfect accuracy on the training set with the traditional approach. However, the significant drop in test accuracy (0.6833) proves that the model is overfitting to the training data. When we use D2C with 2, 3, or 4 subsets, we observe a clear improvement and a much better balance between the training and test accuracies. For all these cases, the training and test accuracies are similar, and the test accuracy for 3 or 4 subsets (0.8333) surpasses that of the traditional approach by a significant margin. This indicates that splitting the data into 3 or 4 subsets provides a good balance between learning patterns from training data and reducing overfitting to improve the model's ability to generalize to unseen data.

Our observations prove that models using the D2C approach perform better on unseen data, as they avoid overfitting by keeping the decision boundary simple and focusing on capturing the broader, more consistent patterns in the data. So, D2C can be seen as yet another technique to reduce overfitting. However, several other techniques address overfitting. So, whether using our proposed method on top of other regularization and augmentation techniques (such as dropouts) improves the performance of a model is the key question at the application level. In the following subsections, we will investigate this using the architectures (incorporating regularization methods like dropouts in the edge models) we described in the previous section on several datasets from different domains. For all these experiments, we used the refined DUA-D2C approach.

\subsection{Analyzing Loss \& Accuracy Curves}
We will analyze the results obtained from our experiments from the 3 different domains. The loss curve is a definitive indicator of overfitting. As we saw in Eq. (\ref{eq5}), The validation/test loss truly represents the \textit{true error} of the model. So we will monitor the validation loss curves from the different experiments we performed to compare and analyze if DUA-D2C improves the generalizing performance, in the form of minimizing the validation loss as the training progresses.\\
\\
\textbf{Image Classification:} Let us first take a look at the loss curves generated using the \textbf{CIFAR-10}, \textbf{Fashion MNIST}, \textbf{MNIST}, and \textbf{FER2013} datasets. For now, we will keep the number of epochs before each round of central aggregation, $\pmb{E}$, equal to $1$. It will ensure a proper comparison between the traditional method and our proposed method while varying the number of subsets, $\pmb{N}$. In our experiments, how much we varied $N$ depended on the performance (Loss curve, Accuracy) trend as we increased $N$, the number of subsets. In the case of all the datasets, we stopped varying $N$ whenever we observed a definitive and consistent performance drop with increasing $\pmb{N}$. For example, in the case of \textbf{CIFAR-10}, we varied $N$ in the range of $1$ to $9$.

\begin{figure}[ht!]
    \centering    
    \includegraphics[width=0.6\columnwidth]{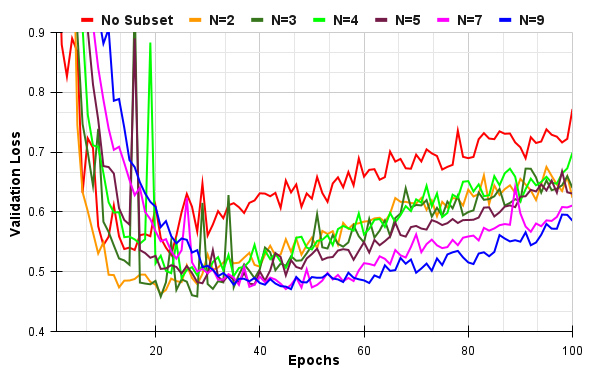}
    \caption{Validation loss curves for the traditional \& the DUA-D2C method for different numbers of subsets, N, on CIFAR-10.}
    \label{fig:cifar-val-loss}
\end{figure}

\begin{figure}[ht!]
        \centering
        \includegraphics[width=0.6\columnwidth]{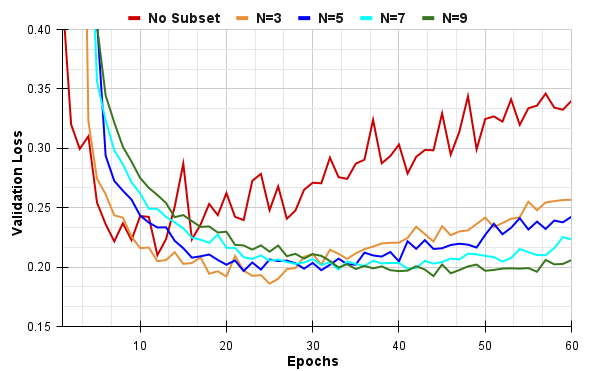}
        \caption{Validation loss curves for the traditional \& the DUA-D2C method for different numbers of subsets, N, on Fashion MNIST.}
        \label{fig:fmnist-val-loss}
\end{figure}

\begin{figure}[ht!]
    \centering
    \includegraphics[width=0.6\columnwidth]{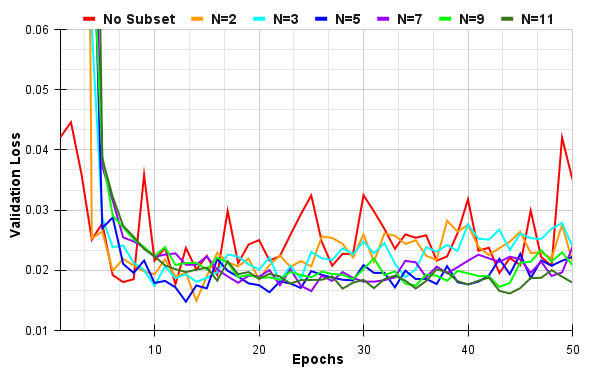}
    \caption{Validation loss curves for the traditional \& the DUA-D2C method for different numbers of subsets, N, on MNIST.}
    \label{fig:mnist-val-loss}
\end{figure}

\begin{figure}[ht!]
    \centering
   \includegraphics[width=0.6\columnwidth]{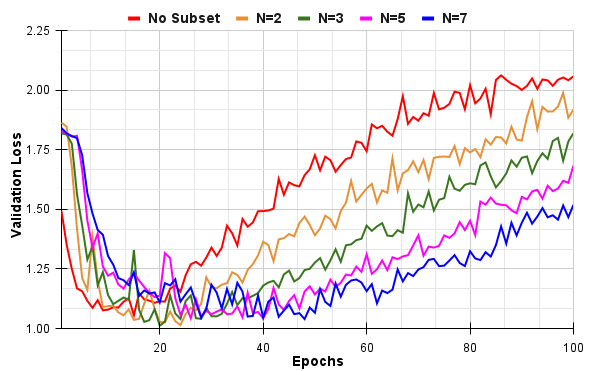}
    \caption{Validation loss curves for the traditional \& the DUA-D2C method for different numbers of subsets, N, on FER2013.}
    \label{fig:fer-val-loss}
\end{figure}

As we can see in Figures \ref{fig:cifar-val-loss}, \ref{fig:fmnist-val-loss}, and \ref{fig:fer-val-loss}, a key indicator of overfitting, an increasing validation loss after initial training, is notably delayed and mitigated when using our proposed DUA-D2C method compared to traditional implementations. In machine learning, overfitting is often marked by a model's increasing validation loss after a certain number of training epochs, indicating that the model has started to capture noise and specific variations in the training set that do not generalize. A model that reaches this overfitting point later, as DUA-D2C does, suggests it remains in a generalization phase for longer. We can also see that, in all these cases, dividing the training data into more subsets with DUA-D2C delays and slows down the abrupt increase in validation loss. For example, on CIFAR-10, the traditional approach (no subsets) reached its minimum loss of $0.536$ at around $16$ epochs, then increased to $0.77$ by 100 epochs. With DUA-D2C ($E=1$), for $N=3$, the minimum loss was $0.462$ around $21$ epochs, rising to $0.64$. When $N=9$, the minimum was $0.47$ around 46 epochs, reaching $0.585$. So, clearly, the convergence to the overfitting point becomes slower, and the abrupt increase of validation loss is minimized as we divide the training set into more subsets, which indicates reduced overfitting with DUA-D2C.

Also, DUA-D2C manages to maintain a lower validation loss in general in all cases, including MNIST, as we can see from Figure \ref{fig:mnist-val-loss}. The minimum value of loss achieved also tells a similar story: we can observe the minimum validation loss for a DUA-D2C variant in each case. When a model maintains a lower validation loss, it indicates better generalization, as it effectively captures true underlying patterns without becoming excessively sensitive to training set noise. This stronger alignment with the data's general structure rather than its idiosyncrasies means the model is overfitting less. The most striking illustration of this phenomenon is seen with the FER2013 dataset (Figure \ref{fig:fer-val-loss}), a task highly prone to overfitting. Here, increasing the number of subsets with DUA-D2C clearly diminishes this overfitting tendency.

This empirical evidence, that employing more subsets helps reduce overfitting, aligns with our theoretical intuition from Eq. (\ref{eq12}). As $k$ (number of subsets) increases, the variance of the aggregated model can decrease (approaching $s/k$ in the ideal scenario), directly combating a primary component of overfitting. Thus, the observed minimization of validation loss (our proxy for true error) with DUA-D2C, particularly with a larger $N$, validates our hypothesis and confirms its effectiveness in treating the problem of overfitting.

There is another interesting aspect of the validation loss curve here. As we can see from Figure \ref{fig:mnist-val-loss}, the validation loss curve generated using the traditional approach deviates much more than the ones generated using the DUA-D2C method. A curve that does not fluctuate much after converging suggests that the model has reached a stable state. This stability is often a sign of robustness, meaning the model is not overly sensitive to slight variations in the data or training process. Dividing the dataset into more and more subsets results in smaller and smaller deviations in the validation loss curve. This was evident in previous loss curves too, to a lesser extent. This proves that dividing the training set into multiple subsets not only improves accuracy and loss but also brings stability and consistency to the loss/accuracy curves.

\begin{figure}[ht!]
        \centering
        \includegraphics[width=0.6\columnwidth]{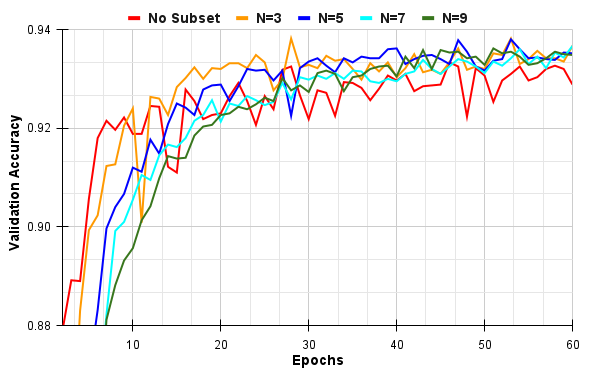}
        \caption{Validation accuracy curves for the traditional \& the DUA-D2C method for different numbers of subsets, N, on Fashion MNIST.}
        \label{fig:fmnist-val-acc}
\end{figure} 

However, the loss curve isn’t the only important factor during the training process. The relation between log loss and accuracy may not be so linear. That is why it is important to observe the validation accuracy too. As we can see from Figure \ref{fig:fmnist-val-acc}, for all the different values of the number of subsets, $N$, even though the loss reaches its minimum much earlier, the accuracy shows an increasing trend (even if slightly) almost throughout the entire training process, right up until the 60th epoch. This shows why continuing training for a significant period might be necessary even if the validation loss starts increasing. This also shows why early stopping based on loss curves may not be a good idea in both cases: the traditional approach and the DUA-D2C method. This is something we observed in the cases of all these datasets. Also, we can notice a visible improvement in terms of validation accuracy in Figure \ref{fig:fmnist-val-acc}. At the very least, dividing the training set into multiple subsets ensures that the maximum accuracy is attained while keeping the log loss relatively in control, which characterizes a more confident and robust model with better generalization ability.

Let us also take a look at the validation loss curves generated using FER2013 while keeping the number of subsets, $N$, constant and varying the number of local epochs before each round of central aggregation, $E$. We varied $E$ between $1$ to $3$ and Kept $N$ constant at $2$ and $3$ separately.

\begin{figure}[ht!]
    \centering
    \includegraphics[width=0.6\columnwidth]{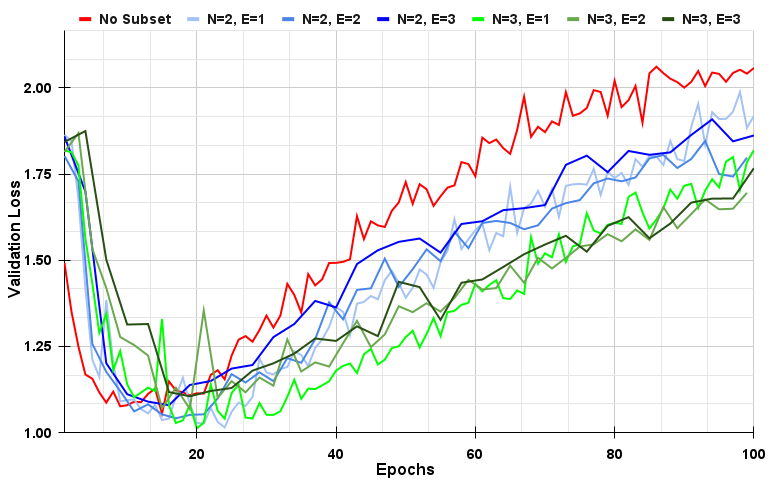}
    \caption{Validation loss curves for different numbers of epochs before each round of central averaging, while keeping the number of subsets, N = 2 \& 3.}
    \label{fig:vary-fer-val-loss}
\end{figure}

We can see from Figure \ref{fig:vary-fer-val-loss}  that when we vary the number of epochs before each round of central aggregation, $E$, there is no clear indication of any kind of change in the trend or the rate of change in the validation losses as the training progresses. This is unlike when we varied the number of subsets, $N$, keeping $E$ constant, where we saw that increasing the number of subsets delayed and minimized the increase in validation loss as the training progressed quite clearly. Also, the convergence got slower. But here, for both $N=2$ (shades of blue) and $N=3$ (shades of green), the trends from the validation curves are quite similar for different values of $E$ throughout the training phase. In light of the above observations, we can conclude that varying the number of epochs before each round of central aggregation, $E$ does not have any obvious implications on log loss as we saw in the case of varying the number of subsets, $N$. However, an increased number of local epochs before each round of central aggregation means more scope for learning from each of the subsets before combining and assigning the weights, sometimes even overfitting on a particular subset a bit more. We will observe the effect of varying the number of epochs further in the later subsections when we observe the final accuracy and log loss in each case. The implication of varying $E$ is likely to be different on different datasets from different domains as we will see in the case of the number of subsets too.

\textbf{Audio Classification:} Overfitting is a common problem in audio classification tasks. But it is more apparent when the model is trained on a particular dataset, and then tested on data from a different source. One way of tackling this issue is by combining datasets to increase the diversity in both training and test/validation data. We combined the datasets TESS \cite{dupuis2010toronto}, CREMA-D \cite{cao2014crema}, and RAVDESS \cite{livingstone2018ryerson} for this task. Let us take a look at the validation loss and accuracy curves generated using this Combined Audio dataset.

\begin{figure}[ht!]
        \centering
        \includegraphics[width=0.6\linewidth]{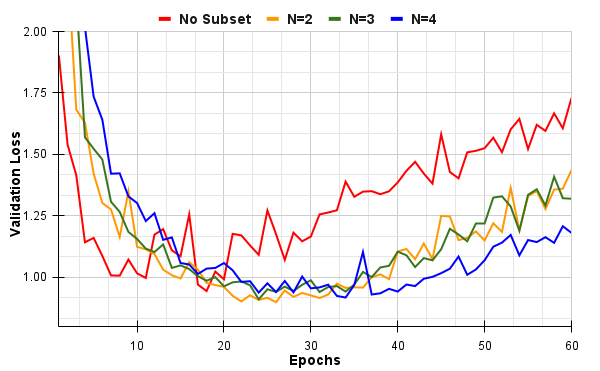}
        \caption{Validation loss curves for the traditional \& the DUA-D2C method for different numbers of subsets, N, on the Combined Audio dataset.}
        \label{fig:audio-val-loss}
\end{figure}

As we can see from Figure \ref{fig:audio-val-loss}, there is overfitting to some extent in all cases, which causes the validation loss to creep up after a few epochs but the rate at which this phenomenon takes place is much less when the training set is divided into multiple subsets. Once again, the rate of increase in validation loss goes down as we increase the number of subsets. Even though we converted the 1D time series audio files to 2D Log-Mel spectrograms to form something like an image, the classification based on these Mel-specs differs greatly from traditional image classification. So, the overfitting-resistant nature of models using our proposed method being apparent in this case is a very promising sign. However, the small size of this dataset may not be ideal in the context of applying DUA-D2C. We will analyze it in detail in the next subsection based on several performance metrics.

\textbf{Text Classification:} Text classification is an important case study when it comes to testing the DUA-D2C method. For both image classification and audio classification, we used 2D data and CNN to carry out the tasks. But text classification deals with 1D sequences, and as we mentioned in the previous section, we used bi-directional LSTM in this case. We used AG News \cite{zhang2015character}, a benchmark dataset to test our method primarily in the NLP domain. Let us observe the validation loss and accuracy curves generated using the AG News dataset first.

\begin{figure}[ht!]
    \centering
    \includegraphics[width=0.6\columnwidth]{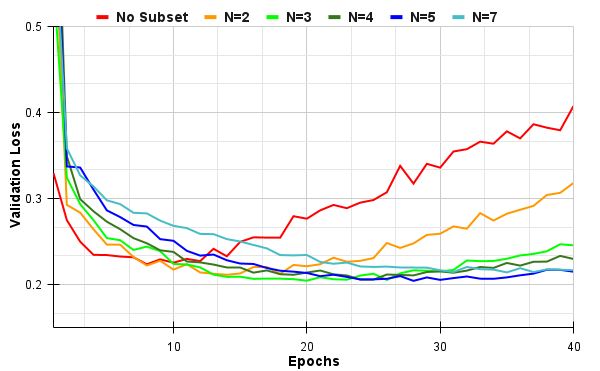}
    \caption{Validation loss curves for the traditional \& the DUA-D2C method for different numbers of subsets on AG News.}
    \label{fig:agn-val-loss}
\end{figure}

\begin{figure}[ht!]
    \centering
    \includegraphics[width=0.6\columnwidth]{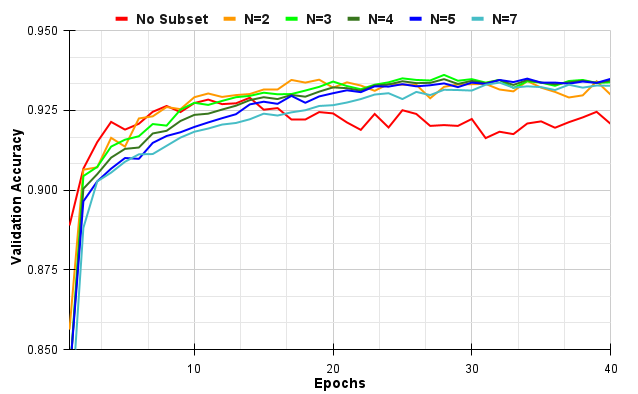}
    \caption{Validation accuracy curves for the traditional \& the DUA-D2C method for different numbers of subsets on AG News.}
    \label{fig:agn-val-acc}
\end{figure}

The loss and accuracy curves generated for AG News depict by far the clearest picture in terms of the improvements brought by the proposed method over the traditional monolithic approach. The loss curve from Figure \ref{fig:agn-val-loss} shows the trend we observed in almost all other cases: dividing the training data into more and more subsets delays and minimizes the increase in validation loss as the training progresses. The convergence though, slows down a bit in this case too. More importantly, with no subsets, the validation loss shoots up sharply after around 8 epochs, which indicates overfitting and impacts the accuracy too as the training progresses as we can see from Figure \ref{fig:agn-val-acc}. However, when we use DUA-D2C, the sharp increase in the validation loss is mitigated drastically. Even with only two subsets, the improvement is quite significant. When the number of subsets is increased further to more than two, the loss curve becomes quite stable, almost completely eliminating the increasing trend and showing consistent improvement as the training progresses. The accuracy curve (Figure \ref{fig:agn-val-acc}) also shows clear indications that DUA-D2C improves over the traditional approach. With the traditional approach, the accuracy too starts deteriorating after around the 14th epoch, indicating severe overfitting, an issue that no longer exists when we use DUA-D2C for any value of $N$ or $E$.

The experiments on all these datasets from 3 different datasets prove that using the DUA-D2C method can indeed address overfitting and control the abrupt increase of true error in the form of maintaining a lower validation loss during training. This also shows that DUA-D2C works well in different domains and a diverse set of tasks.

\subsection{Accuracy, F1 Score, AUC-ROC, Cohen's Cappa, Log Loss and MCC}
We evaluated our proposed DUA-D2C on test sets using Accuracy, F1 Score, AUC-ROC, Cohen's Cappa, Log Loss, and Matthew's Correlation Coefficient (MCC) to capture a comprehensive view of model performance in terms of correctness, class balance, prediction confidence, and discrimination ability. Reporting these final metrics on test sets was essential to confirm that improvements generalized well beyond the training and validation data, verifying true effectiveness in unseen scenarios. We used the average from multiple repetitive runs (to test for repeatability) for each of these metrics. We used the provided test sets with each of the datasets for proper comparisons. In the tables, the top row represents the traditional centralized approach, and the rest of the rows represent different variations of the DUA-D2C approach. The bold entities represent the best performance in the tables.

\textbf{Image Classification:} Let’s take a look at the performance metrics from the benchmark image classification datasets we primarily used using different settings (the traditional approach and the DUA-D2C method with different values of $N$ and $E$).

\begin{table}[ht]
    \renewcommand{\arraystretch}{1.2}
    \centering
    \caption{\label{tab:cifar-tab} Performance Metrics with Traditional Approach vs. Different Settings Using the DUA-D2C Method on CIFAR-10.}
    \begin{tabular}{|c|c|c|c|c|c|c|}
    \hline
    \textit{Model Settings} & \textit{Accuracy} & \textit{F1-Score} & \textit{ROC-AUC} & \textit{Cohen Kappa} & \textit{Log Loss} & \textit{Matthews CC} \\ \hline \hline
    No Subsets              & 0.8604            & 0.8601            & 0.98731          & 0.8458             & 0.7198            & 0.8458             \\ 
    N=2, E=1                & 0.8626            & 0.8628            & 0.98704          & 0.8476             & 0.7002            & 0.8476             \\ 
    N=3, E=1                & \textbf{0.8652}   & \textbf{0.8648}   & \textbf{0.98761} & \textbf{0.8501}    & 0.6491            & \textbf{0.8501}    \\ 
    N=3, E=2                & 0.8630            & 0.8624            & 0.98719          & 0.8478             & 0.6215            & 0.8478             \\ 
    N=4, E=1                & 0.8595            & 0.8591            & 0.98710          & 0.8439             & 0.6416            & 0.8439             \\ 
    N=5, E=1                & 0.8614            & 0.8613            & 0.98676          & 0.8461             & 0.6347            & 0.8461             \\ 
    N=7, E=1                & 0.8556            & 0.8553            & 0.98658          & 0.8393             & 0.6126            & 0.8398             \\ 
    N=9, E=1                & 0.8548            & 0.8542            & 0.98664          & 0.8385             & \textbf{0.5876}   & 0.8390             \\ \hline
    \end{tabular}
\end{table}

\begin{table}[ht]
    \renewcommand{\arraystretch}{1.2}
    \centering
    \caption{\label{tab:fashion-mnist-tab} Performance Metrics with Traditional Approach vs. Different Settings Using the DUA-D2C Method on Fashion MNIST.}
    \begin{tabular}{|c|c|c|c|c|c|c|}
    \hline
    \textit{Model Settings} & \textit{Accuracy} & \textit{F1-Score} & \textit{ROC-AUC} & \textit{Cohen Kappa} & \textit{Log Loss} & \textit{Matthews CC} \\ \hline \hline
    No Subsets              & 0.9312            & 0.9313            & 0.99546          & 0.9237             & 0.3492            & 0.9238             \\ 
    N=3, E=1                & 0.9338            & 0.9336            & 0.99570          & 0.9264             & 0.2865            & 0.9265             \\ 
    N=3, E=2                & \textbf{0.9344}   & \textbf{0.9340}   & 0.99571          & \textbf{0.9271}    & 0.3006            & \textbf{0.9272}    \\ 
    N=3, E=3                & 0.9332            & 0.9329            & 0.99580          & 0.9258             & 0.3008            & 0.9258             \\ 
    N=5, E=1                & 0.9327            & 0.9323            & 0.99581          & 0.9252             & 0.2618            & 0.9253             \\ 
    N=7, E=1                & 0.9321            & 0.9321            & 0.99591          & 0.9246             & 0.2383            & 0.9246             \\ 
    N=7, E=2                & 0.9322            & 0.9319            & \textbf{0.99600} & 0.9247             & 0.2377            & 0.9247             \\ 
    N=9, E=1                & 0.9300            & 0.9299            & 0.99590          & 0.9222             & \textbf{0.2314}   & 0.9222             \\ \hline
    \end{tabular}
\end{table}

\begin{table}[ht]
    \renewcommand{\arraystretch}{1.2}
    \centering
    \caption{\label{tab:mnist-tab} Performance Metrics with Traditional Approach vs. Different Settings Using the DUA-D2C Method on MNIST.}
    \begin{tabular}{|c|c|c|c|c|c|c|}
    \hline
    \textit{Model Settings} & \textit{Accuracy} & \textit{F1-Score} & \textit{ROC-AUC} & \textit{Cohen Kappa} & \textit{Log Loss} & \textit{Matthews CC} \\ \hline \hline
    No Subsets              & 0.9952            & 0.9952            & 0.99995          & 0.9947             & 0.0258            & 0.9947             \\ 
    N=2, E=1                & 0.9954            & 0.9954            & 0.99991          & 0.9949             & 0.0294            & 0.9949             \\ 
    N=3, E=1                & 0.9955            & 0.9955            & 0.99995          & 0.9950             & 0.0225            & 0.9950             \\ 
    N=3, E=2                & 0.9953            & 0.9953            & 0.99995          & 0.9948             & 0.0259            & 0.9948             \\ 
    N=5, E=1                & 0.9957            & 0.9957            & 0.99998          & 0.9952             & 0.0188            & 0.9952             \\ 
    N=7, E=1                & 0.9960            & 0.9959            & 0.99998          & 0.9956             & 0.0183            & 0.9956             \\ 
    N=9, E=1                & \textbf{0.9963}   & \textbf{0.9963}   & \textbf{0.99999} & \textbf{0.9959}    & \textbf{0.0148}   & \textbf{0.9959}    \\ 
    N=11, E=1               & 0.9953            & 0.9953            & 0.99998          & 0.9948             & 0.0158            & 0.9948             \\ 
    N=11, E=2               & 0.9959            & 0.9959            & 0.99998          & 0.9954             & 0.0156            & 0.9954             \\ \hline
    \end{tabular}
\end{table}

As we can see from Table \ref{tab:cifar-tab}, Table \ref{tab:fashion-mnist-tab}, and Table \ref{tab:mnist-tab}, the best accuracy, F1-score, MCC, Cohen's Cappa, log loss, and AUC-ROC were achieved using the DUA-D2C method for all three datasets. For \textbf{CIFAR-10}, the number of subsets, $N$, was $3$ for the best model in terms of accuracy. The number of epochs before each round of central aggregation, $E$, was just $1$. This approach showed an improvement with an accuracy of $86.52\%$ over the traditional approach with no subsets, which yielded an accuracy of $86.04\%$. Initially, the accuracy improved as we increased the number of subsets, but suffered a decreasing trend after it was further increased past $3$. As we discussed, dividing the training set into too many subsets may impact training negatively due to a lack of representative samples. The experimental outcomes show further evidence of that in this case.

For \textbf{Fashion MNIST} too, the best metrics were achieved using our proposed DUA-D2C method. The number of subsets, $N$, was once again $3$ in this case for the best model in terms of accuracy. The number of epochs before each round of central aggregation, $E$, was $2$ in this case. This approach again showed an improvement with an accuracy of $93.44\%$ over the traditional approach with no subsets, which yielded an accuracy of $93.12\%$. An increasing trend is visible in ROC-AUC as we increase the number of subsets till it reaches the best conditions with $N = 7$. 

In the case of \textbf{MNIST} too, the best metrics were all achieved using our proposed method. The number of subsets, $N$, was $9$ in this case for the best model. The number of epochs before each round of central aggregation, $E$, was $1$. The improvement of $0.11$ in error rate over the traditional approach is very significant for the \textbf{MNIST} dataset. In this case, the least amount of log loss is also achieved with the best model in terms of accuracy.

We can also see that as we increase the number of subsets, $N$, there is a clear trend of decreasing log loss. As we discussed earlier, increasing the number of subsets further minimizes the effect of noise and outliers. The decrease in log loss is a result of that phenomenon. Even though the best model isn’t usually selected based on log loss values, in the case of similar accuracy, the best model can be selected based on log loss. Log Loss evaluates how well the model predicts probabilities, penalizing confident but incorrect predictions, which is critical for understanding model reliability. A lower log loss means a more robust and confident model, so the model with a lower log loss is preferred when the test accuracy or F1 score is similar.

The comparison of subset numbers in the best-performing models across \textbf{CIFAR-10}, \textbf{Fashion MNIST}, and \textbf{MNIST} provides critical insights. For CIFAR-10, the optimal model utilized only $3$ training subsets, while the best models for Fashion MNIST and MNIST required $3$ ($7$ in terms of ROC-AUC) and $9$ subsets, respectively. Notably, increasing the subset count beyond $3$ led to significantly declining accuracy for CIFAR-10, a trend not observed with the other datasets. This discrepancy is likely due to the distinct characteristics of each dataset. While all contain $10$ classes, CIFAR-10, a smaller dataset with RGB images and complex class representations, presents a more challenging classification task compared to the grayscale images of Fashion MNIST and MNIST. Consequently, effective training on CIFAR-10 demands larger sample sizes per subset, and when these subsets become too small, performance degrades. In contrast, Fashion MNIST and MNIST, identical in dataset size, differ in task complexity, with Fashion MNIST presenting a slightly more challenging classification task than MNIST. This complexity results in the optimal Fashion MNIST model requiring fewer subsets than MNIST, as each subset’s sample size needs to be sufficient for adequate training.

This analysis reiterates a key principle for applying the DUA-D2C method effectively: to divide training data into more subsets, datasets must be large enough to allow each subset enough training data. Larger datasets benefit more from divided training, requiring more subsets to achieve optimal performance.

\begin{table}[ht]
    \renewcommand{\arraystretch}{1.2}
    \centering
    \caption{\label{tab:fer-tab} Performance Metrics with Traditional Approach vs. Different Settings Using the DUA-D2C Method on FER2013.}
    \begin{tabular}{|c|c|c|c|c|c|c|}
    \hline
    \textit{Model Settings} & \textit{Accuracy} & \textit{F1-Score} & \textit{ROC-AUC} & \textit{Cohen Kappa} & \textit{Log Loss} & \textit{Matthews CC} \\ \hline \hline
    No Subsets              & 0.6498            & 0.6420            & 0.89934          & 0.5756             & 2.0112            & 0.5759             \\ 
    N=2, E=1                & 0.6512            & \textbf{0.6457}   & 0.90262          & 0.5781             & 1.6723            & 0.5783             \\ 
    N=2, E=2                & 0.6510            & 0.6381            & 0.90360          & 0.5776             & 1.5595            & 0.5777             \\ 
    N=2, E=3                & 0.6480            & 0.6338            & 0.89610          & 0.5740             & 1.6505            & 0.5750             \\ 
    N=3, E=1                & 0.6471            & 0.6365            & 0.89910          & 0.5735             & 1.6408            & 0.5739             \\ 
    N=3, E=2                & 0.6482            & 0.6346            & 0.90088          & 0.5743             & 1.6311            & 0.5749             \\ 
    N=3, E=3                & 0.6420            & 0.6232            & 0.89670          & 0.5671             & 1.4744            & 0.5682             \\ 
    N=5, E=1                & \textbf{0.6537}   & 0.6370            & \textbf{0.90528} & \textbf{0.5816}    & \textbf{1.1934}   & \textbf{0.5818}    \\ 
    N=5, E=2                & 0.6482            & 0.6308            & 0.90020          & 0.5741             & 1.4750            & 0.5747             \\ 
    N=7, E=1                & 0.6474            & 0.6339            & 0.90244          & 0.5732             & 1.3299            & 0.5743             \\ \hline
    \end{tabular}
\end{table}

FER2013, a challenging dataset for facial expression recognition, represents a task that is particularly prone to overfitting. As seen in Table \ref{tab:fer-tab}, our proposed method outperformed the traditional approach and achieved the best accuracy, Cohen's Cappa, MCC, log loss, and AUC-ROC, with the best-performing model using 5 subsets ($N=5$) and 1 training epoch before each central aggregation step ($E=1$). DUA-D2C yielded a notable accuracy improvement, reaching 65.37\% compared to 64.98\% of the traditional model without subsets. However, D2C performed even better with up to 65.67\% of accuracy using the best settings. A decreasing trend in log loss was observed with more subsets, though no consistent pattern emerged with varying epochs ($E$).

FER2013 is a particularly important test case for us as this dataset represents a comparatively \textbf{imbalanced class distribution}. Due to FER2013's class imbalance, accuracy alone may not reliably indicate the best model, so, the F1-score, a metric considering class-wise precision, provides a valuable perspective. Interestingly, the optimal F1-score model (2 subsets, $E=1$) differs from the accuracy-optimized model, with D2C achieving an F1 score of \textbf{0.6457} versus \textbf{0.642} in the baseline model. This F1-score improvement underscores DUA-D2C’s effectiveness in addressing overfitting, particularly in class-imbalanced, complex tasks.

\textbf{Audio Classification:} Now let us take a look at the performance metrics from the Combined Audio dataset. This is another important case as combining datasets like TESS, CREMA-D, and RAVDESS is likely to lead to Non-IID data, and this one is another imbalanced and smaller dataset.

\begin{table}[ht]
    \renewcommand{\arraystretch}{1.2}
    \centering
    \caption{\label{tab:audio-combined-tab} Performance Metrics with Traditional Approach vs. Different Settings Using DUA-D2C on the Combined Audio Dataset.}
    \begin{tabular}{|c|c|c|c|c|c|c|}
    \hline
    \textit{Model Settings} & \textit{Accuracy} & \textit{F1-Score} & \textit{ROC-AUC} & \textit{Cohen Kappa} & \textit{Log Loss} & \textit{Matthews CC} \\ \hline \hline
    No Subsets              & 0.6953            & 0.6870            & \textbf{0.96570} & 0.6684             & 1.3419            & 0.6692             \\ 
    N=2, E=1                & 0.6869            & 0.6787            & 0.96226          & 0.6592             & 1.7225            & 0.6601             \\ 
    N=2, E=2                & \textbf{0.7010}   & \textbf{0.6973}   & 0.96360          & \textbf{0.6747}    & 1.5021            & \textbf{0.6751}    \\ 
    N=3, E=1                & 0.6946            & 0.6849            & 0.96284          & 0.6678             & 1.4576            & 0.6681             \\ 
    N=3, E=2                & 0.6775            & 0.6695            & 0.96560          & 0.6492             & \textbf{1.0571}   & 0.6504             \\ 
    N=4, E=1                & 0.6930            & 0.6805            & 0.96423          & 0.6659             & 1.3829            & 0.6662             \\ 
    N=5, E=1                & 0.6938            & 0.6862            & 0.96523          & 0.6668             & 1.2160            & 0.6673             \\ \hline
    \end{tabular}
\end{table}

We can see from Table \ref{tab:audio-combined-tab} that in the \textbf{Combined Audio} dataset, the best metrics were achieved using our proposed DUA-D2C method once again. The traditional model attained a 69.53\% accuracy, while the DUA-D2C approach, reached 70.10\% with $2$ subsets and $E =2$. Accuracy declined as the number of subsets increased, likely due to the smaller dataset size (11,632 samples), limiting effective data availability per subset. As a result, dividing the dataset into smaller portions likely impacted performance by reducing the sample count for each class, a significant challenge given the dataset's complexity and diversity, as it combines samples from multiple sources.

However, despite declining accuracy, the DUA-D2C method showed reduced log loss as $N$ increased to 3, indicating a trade-off between accuracy and confidence in predictions. Given the dataset’s class imbalance, the F1-score is a more reliable metric here. The DUA-D2C approach yielded a significant improvement in the F1-score (0.6973 vs. 0.687), indicating better generalization across classes. This suggests that DUA-D2C supports robust classification with fewer, more distinctive errors, especially in class-imbalanced, complex tasks like this one. However, the traditional approach outperformed both DUA-D2C and D2C in terms of ROC-AUC for this dataset. In summary, while DUA-D2C did not maximize ROC-AUC, it achieved better metrics overall, resulting in a model that generalizes well and offers more confident predictions in challenging, class-imbalanced datasets. However, dividing the dataset into too many subsets does worsen the performance of the models in this case.

This analysis leads us to the conclusion that \textbf{to apply the DUA-D2C method effectively, the subsets need to be sufficiently large so that each of them can have enough training data}. It is however noteworthy that even with a comparatively smaller dataset, DUA-D2C still shows performance improvement in most metrics.

\textbf{Text Classification:} Finally, let us take a look at the performance metrics from the AG News dataset. 

\begin{table}[ht]
    \renewcommand{\arraystretch}{1.2}
    \centering
    \caption{\label{tab:ag-news-tab} Performance Metrics with Traditional Approach vs. Different Settings Using the DUA-D2C Method on AG News.}
    \begin{tabular}{|c|c|c|c|c|c|c|}
    \hline
    \textit{Model Settings} & \textit{Accuracy} & \textit{F1-Score} & \textit{ROC-AUC} & \textit{Cohen Kappa} & \textit{Log Loss} & \textit{Matthews CC} \\ \hline \hline
    No Subsets              & 0.9243            & 0.9243            & 0.98764          & 0.8991             & 0.2611            & 0.8992             \\ 
    N=2, E=1                & 0.9300            & 0.9300            & 0.98856          & 0.9067             & 0.2519            & 0.9067             \\ 
    N=3, E=1                & \textbf{0.9312}   & \textbf{0.9312}   & 0.98888          & \textbf{0.9082}    & \textbf{0.2378}   & \textbf{0.9084}    \\ 
    N=3, E=2                & 0.9278            & 0.9277            & 0.98879          & 0.9037             & 0.2411            & 0.9037             \\ 
    N=5, E=1                & 0.9274            & 0.9273            & \textbf{0.98905} & 0.9032             & 0.2416            & 0.9032             \\ 
    N=5, E=2                & 0.9287            & 0.9287            & 0.98830          & 0.9049             & 0.2656            & 0.9049             \\ 
    N=7, E=1                & 0.9288            & 0.9288            & 0.98836          & 0.9051             & 0.2443            & 0.9052             \\ \hline
    \end{tabular}
\end{table}

We can observe from Table \ref{tab:ag-news-tab} that, once again, the highest accuracy and F1-score were achieved with our proposed DUA-D2C method. The number of subsets, $N$, was $3$ in this case for the best model. The number of epochs before each round of central averaging, $E$, was $1$. The DUA-D2C method showed a significant improvement with an accuracy of $93.12\%$ over the traditional approach with no subsets, which yielded an accuracy of $92.43\%$. The best method in terms of AUC-ROC happens to be the model with even more subsets (\textbf{5}). It also seems that the log loss increases a bit when we increase the number of epochs, $E$.

The effect of using DUA-D2C is the most apparent in this dataset compared to all the previous ones. AG News, the largest dataset in our study, presents an ideal case for DUA-D2C, as each class contains 30,000 samples. This abundance of data per class allows for substantial division without sacrificing subset size, which maintains the model’s training stability while leveraging the regularizing effect of partitioning and averaging. Consequently, every variation of DUA-D2C (across all values of $N$ and $E$) demonstrated accuracy gains over the traditional approach, underscoring DUA-D2C’s suitability for mitigating overfitting in text classification tasks.

In summary, DUA-D2C excels with large datasets like AG News, where ample samples per class allow the model to benefit from both training on robust subsets and reducing overfitting through partitioning and averaging. This validates DUA-D2C’s potential for significant impact in Big Data applications, where traditional models often struggle to fully exploit available data.


\begin{table}[t]
  \caption{Comparison of Best Performance: Traditional vs. D2C vs. DUA-D2C across Datasets for Key Metrics.}
  \label{tab:comparison-summary}
  \begin{threeparttable}
    \renewcommand{\arraystretch}{1.3}
    \centering
    \begin{tabular}{@{}l l c c c@{}}
      \toprule
      \textit{Dataset} & \textit{Metric} & \textit{Traditional} & \textit{D2C (Best)} & \textit{DUA-D2C (Best)} \\
      \midrule\midrule

      \multirow{4}{*}{CIFAR-10}
      & Accuracy   & 0.8604 & 0.8613 & \textbf{0.8652} \\
      & F1-Score   & 0.8601 & 0.8611 & \textbf{0.8648} \\
      & ROC-AUC    & 0.98731 & 0.98704 & \textbf{0.98761} \\
      & Log Loss   & 0.7198 & \textbf{0.5733} & 0.5876 \\
      \midrule

      \multirow{4}{*}{Fashion MNIST}
      & Accuracy   & 0.9312 & 0.9338 & \textbf{0.9344} \\
      & F1-Score   & 0.9313 & 0.9337 & \textbf{0.9340} \\
      & ROC-AUC    & 0.99546 & \textbf{0.99600} & \textbf{0.99600} \\
      & Log Loss   & 0.3492 & 0.2341 & \textbf{0.2314} \\
      \midrule

      \multirow{4}{*}{MNIST}
      & Accuracy   & 0.9952 & 0.9959 & \textbf{0.9963} \\
      & F1-Score   & 0.9952 & 0.9959 & \textbf{0.9963} \\
      & ROC-AUC    & 0.99995 & \textbf{0.99999} & \textbf{0.99999} \\
      & Log Loss   & 0.0258 & 0.0161 & \textbf{0.0148} \\
      \midrule

      \multirow{4}{*}{FER2013}
      & Accuracy   & 0.6498 & \textbf{0.6567} & 0.6537 \\
      & F1-Score   & 0.6420 & \textbf{0.6465} & 0.6457 \\
      & ROC-AUC    & 0.89934 & 0.90320 & \textbf{0.90528} \\
      & Log Loss   & 2.0112 & 1.3941 & \textbf{1.1934} \\
      \midrule

      \multirow{4}{*}{Audio\_Combined}
      & Accuracy   & 0.6953 & 0.6942 & \textbf{0.7010} \\
      & F1-Score   & 0.6870 & 0.6909 & \textbf{0.6973} \\
      & ROC-AUC    & \textbf{0.96570} & 0.96553 & 0.96560 \\
      & Log Loss   & 1.3419 & \textbf{1.0033} & 1.0571 \\
      \midrule

      \multirow{4}{*}{AG News}
      & Accuracy   & 0.9243 & 0.9295 & \textbf{0.9312} \\
      & F1-Score   & 0.9243 & 0.9296 & \textbf{0.9312} \\
      & ROC-AUC    & 0.98764 & \textbf{0.98916} & 0.98905 \\
      & Log Loss   & 0.2611 & \textbf{0.2353} & 0.2378 \\
      \bottomrule
    \end{tabular}

    \begin{tablenotes}[para,flushleft]
      \footnotesize
      \item Values in D2C (Best) and DUA-D2C (Best) columns represent the optimal performance across all tested N (Number of Subsets) and E (Local Epochs) configurations for that method. Bold indicates the best overall performance among the three approaches (Traditional, D2C, DUA-D2C) for each metric and dataset. For Log Loss, lower values are better.
    \end{tablenotes}
  \end{threeparttable}
\end{table}

The results presented in Table \ref{tab:comparison-summary} demonstrates the advantages of the "divide and conquer" strategy. The D2C method consistently outperforms the traditional monolithic training approach across various datasets and key metrics, underscoring the inherent benefits of partitioned training and parameter averaging for improved generalization.

Building upon this foundation, the DUA-D2C method further refines and perfects this technique. Through its dynamic, uncertainty-aware aggregation, DUA-D2C consistently achieves superior performance compared to both the traditional method and the standard D2C. While the D2C method itself provides substantial gains, the intelligent weighting scheme of DUA-D2C offers an additional performance enhancement. This positions DUA-D2C as the more advanced and robust implementation of our proposed framework, consistently delivering strong, and in many cases, the best, generalization capabilities.

\subsection{Augmentation in DUA-D2C}
Our experiments show that applying DUA-D2C results in less overfitting and an improvement in performance metrics like accuracy, F1 score, etc. However, we also saw that the training subsets need to be sufficiently large so that each of the subsets can provide enough training diversity to the edge models, otherwise, the performance suffers a drop. When the dataset is large, the proposed method works very well, but when the dataset is smaller, this may cause issues in achieving optimum performance.

One way to counter these issues is data augmentation. This is a widely used technique to expand a dataset and prevent overfitting. To minimize the effect of outliers and noise and to prevent data leakage among the subsets, augmentation must take place \textbf{after} dividing the dataset into multiple subsets. The augmentation is then applied to each of the subsets. That way, the augmented samples from one subset won’t get mixed with another subset. This approach ensures the separation of data in each of the subsets as well as the expansion of training data, providing enough samples for all the subsets, thus improving the overall performance. We tested this augmentation approach on the \textbf{CIFAR-10} dataset for $3$, $5$ \& $7$ subsets. The augmentation methods we used were simple rotation, horizontal flip, and shifting along both axes. Let us take a look at the accuracies we achieved in each case and compare them with the accuracies we got before without applying augmentation. 

\begin{figure}[ht!]
\centering
\includegraphics[width=0.7\linewidth]{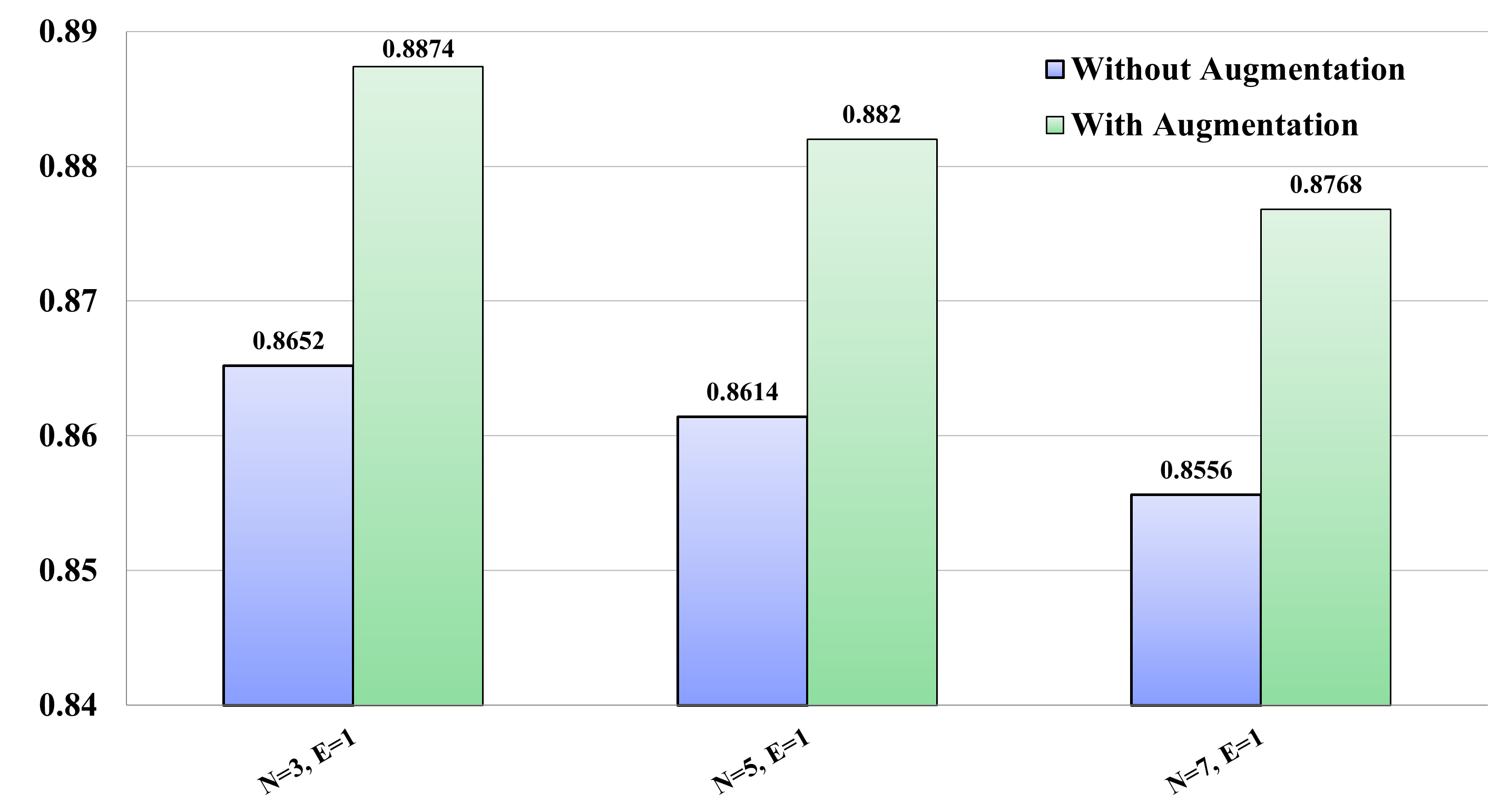}
\caption{Test accuracies for different numbers of subsets, N, using the DUA-D2C method on the CIFAR-10 dataset with and without Augmentation.}
\label{fig:augment-acc-bar}
\end{figure}

As we can see from Figure \ref{fig:augment-acc-bar}, the accuracy improved significantly in each case after we applied data augmentation to the subsets of the training data. The accuracy increased from $86.52\%$ to $88.74\%$ for $3$ subsets. In the case of $5$ subsets, the accuracy reached $88.12\%$ from just $86.14\%$. In the case of $7$ subsets too, the accuracy increased from $85.56\%$ to $87.68\%$. This shows that data augmentation techniques can be successfully applied when using the DUA-D2C method, and they can effectively treat the problems that arise when the size of the subsets becomes too small.

\subsection{Sensitivity Analysis and Ablation Studies}
To rigorously validate the robustness of the proposed framework and verify that the performance gains are not artifacts of hyperparameter tuning or standard regularization techniques, we conducted two targeted ablation studies.

\subsubsection{Orthogonality to Standard Regularization}
A critical question regarding any new regularization technique is whether its benefits are redundant when standard internal regularizers (such as Dropout and Batch Normalization) are already applied. To address this, we compared the performance of the baseline  against DUA-D2C (using the optimal $N$ for each dataset) under two distinct configurations:
\begin{enumerate}
    \item \textbf{No Regularization:} Removing all Dropout and Batch Normalization layers from the architecture.
    \item \textbf{With Regularization:} The standard architecture including Dropout layers and Batch Normalization.
\end{enumerate}
We performed this analysis on three diverse datasets: \textbf{CIFAR-10} (Image), \textbf{Combined Audio Dataset} (Audio, Imbalanced), and \textbf{AG News} (Text). The results are detailed in Table \ref{tab:ablation-combined}.

\begin{table}[ht]
    \renewcommand{\arraystretch}{1.2}
    \setlength{\tabcolsep}{3pt} 
    \centering
    \small 
    \caption{\label{tab:ablation-combined} Impact of DUA-D2C with and without Internal Regularization across diverse datasets.}
    \begin{tabular}{|l|l|c|c|c|c|}
    \hline
    \textbf{Dataset} & \textbf{Configuration} & \textbf{Accuracy} & \textbf{F1-Score} & \textbf{ROC-AUC} & \textbf{Log Loss} \\ \hline \hline
    
    \multirow{4}{*}{\textbf{CIFAR-10}} 
    & Baseline (No Reg) & 0.7192 & 0.7196 & 0.9556 & 2.4839 \\ 
    & \textbf{DUA-D2C (No Reg)} & \textbf{0.7397} & \textbf{0.7380} & \textbf{0.9601} & \textbf{0.9092} \\
    & Baseline (With Reg) & 0.8612 & 0.8608 & 0.9873 & 0.7198 \\ 
    & \textbf{DUA-D2C (With Reg)} & \textbf{0.8652} & \textbf{0.8648} & \textbf{0.9876} & \textbf{0.5876} \\ \hline

    \multirow{4}{*}{\shortstack[l]{\textbf{Combined}\\\textbf{Audio}}} 
    & Baseline (No Reg) & 0.6125 & 0.6056 & 0.9442 & 3.1117 \\ 
    & \textbf{DUA-D2C (No Reg)} & \textbf{0.6210} & \textbf{0.6062} & \textbf{0.9500} & \textbf{1.2036} \\
    & Baseline (With Reg) & 0.6953 & 0.6870 & 0.9657 & 1.3419 \\ 
    & \textbf{DUA-D2C (With Reg)} & \textbf{0.7010} & \textbf{0.6973} & \textbf{0.9657} & \textbf{1.0571} \\ \hline

    \multirow{4}{*}{\textbf{AG News}} 
    & Baseline (No Reg) & 0.9267 & 0.9266 & 0.9884 & 0.2975 \\ 
    & \textbf{DUA-D2C (No Reg)} & \textbf{0.9309} & \textbf{0.9310} & \textbf{0.9889} & \textbf{0.2380} \\
    & Baseline (With Reg) & 0.9243 & 0.9243 & 0.9876 & 0.2611 \\ 
    & \textbf{DUA-D2C (With Reg)} & \textbf{0.9312} & \textbf{0.9312} & \textbf{0.9890} & \textbf{0.2378} \\ \hline
    \end{tabular}
\end{table}

\textbf{General Observation (CIFAR-10):} DUA-D2C yields consistent performance improvements in both the presence and absence of internal regularization. On CIFAR-10 (Table \ref{tab:ablation-combined}), removing internal regularizers causes a drastic drop in the Baseline performance (accuracy falls from 0.8612 to 0.7192). However, applying DUA-D2C to these unregularized models results in an improved performance (+2.05\% accuracy). This proves again that the ``divide to conquer" data partitioning strategy acts as a powerful structural regularizer in its own right, compensating for the lack of Dropout.

\textbf{Observation on Orthogonality (AG News):} Strikingly, for the AG News dataset (Table \ref{tab:ablation-combined}), the unregularized Baseline (0.9267) actually outperformed the regularized Baseline (0.9243). This suggests that for this specific text classification architecture, standard microscopic regularizers like Dropout were detrimental. Crucially, DUA-D2C provided clear improvements in \textbf{both} scenarios, pushing the accuracy to $\approx 0.931$ and lowering the Log Loss. This indicates that while internal regularizers failed, the macroscopic regularization introduced by DUA-D2C succeeded. It confirms that DUA-D2C is \textbf{orthogonal} to standard techniques, providing additive gains when they work, and corrective gains when they do not.

\textbf{Impact on Imbalanced Data (Audio):} The results on the Audio dataset (Table \ref{tab:ablation-combined}) highlight the specific strength of DUA-D2C in handling imbalance and uncertainty. Once again, DUA-D2C brings improvement both in terms of accuracy and F1-score in all cases, even though the improvement is comparatively minimal due to the smaller subset size. Also, in the ``No Regularization" case, the Baseline model exhibits an extremely high Log Loss of 3.1117 despite achieving moderate accuracy. This indicates severe poor calibration: the model is making ``confidently wrong" predictions, likely overfitting to the majority class or noise. In contrast, DUA-D2C not only improves the accurany by nearly 1\% but also drastically reduces this Log Loss to 1.2036. This massive reduction confirms that our uncertainty-aware aggregation (utilizing Entropy) effectively identifies and down-weights these overconfident, uncalibrated edge models, resulting in a central model that is not only more accurate but significantly more reliable and robust to imbalance.

\subsubsection{Sensitivity to Tradeoff Parameter ($\lambda$)}
The hyperparameter $\lambda$ governs the balance between Validation Accuracy and Inverse Entropy in the aggregation logic (Eq. \ref{eq:composite_score}). To determine the necessity of the hybrid approach, we evaluated DUA-D2C on \textbf{CIFAR-10} while varying $\lambda$ from $0.0$ (Entropy only) to $1.0$ (Accuracy only).

\begin{table}[ht]
    \renewcommand{\arraystretch}{1.2}
    \centering
    \small
    \caption{\label{tab:lambda-sensitivity} Sensitivity Analysis of Tradeoff Parameter $\lambda$ on CIFAR-10 Metrics.}
    \begin{tabular}{|c|c|c|c|c|c|}
    \hline
    \textbf{$\lambda$} & \textbf{Focus} & \textbf{Acc.} & \textbf{F1} & \textbf{AUC} & \textbf{Log Loss} \\ \hline \hline
    0.0 & Entropy Only    & 0.8624 & 0.8619 & 0.9868 & 0.6850 \\
    0.2 & Mostly Entropy  & 0.8628 & 0.8630 & 0.9864 & 0.6921 \\
    0.5 & Balanced        & 0.8640 & 0.8638 & 0.9874 & 0.6543 \\
    \textbf{0.7} & \textbf{Hybrid} & \textbf{0.8652} & \textbf{0.8648} & \textbf{0.9876} & 0.6491 \\
    0.8 & Mostly Acc.     & 0.8642 & 0.8641 & 0.9873 & \textbf{0.6312} \\
    1.0 & Accuracy Only   & 0.8644 & 0.8640 & 0.9871 & 0.6328 \\ \hline
    \end{tabular}
\end{table}

As detailed in Table \ref{tab:lambda-sensitivity}, the results provide a compelling justification for the hybrid weighting scheme:

\begin{itemize}
    \item \textbf{Failure of Uncertainty Alone ($\lambda \to 0$):} Relying solely on entropy ($\lambda=0.0$) yields the lowest performance across all metrics. This highlights the danger of ``blind confidence'': edge models may be highly certain about their predictions (low entropy) while being factually incorrect. Without the corrective signal of validation accuracy, these confident-but-wrong models corrupt the central aggregation.
    
    \item \textbf{Sub-optimality of Accuracy Alone ($\lambda \to 1$):} Crucially, ignoring uncertainty entirely ($\lambda=1.0$) results in suboptimal performance for \textit{both} classification and calibration. The Log Loss at $\lambda=1.0$ (0.6328) is higher than at $\lambda=0.8$ (0.6312). This indicates that validation accuracy alone is an insufficient proxy for model quality; incorporating even a small uncertainty penalty (as in $\lambda=0.8$) helps filter out overfitted models that are accurate but poorly calibrated.
    
    \item \textbf{The Hybrid Sweet Spot ($\lambda=0.7$):} The peak performance for Accuracy, F1, and ROC-AUC is consistently observed at $\lambda=0.7$. While $\lambda=0.8$ offers slightly better Log Loss, the 0.7 weighting provides the maximum discriminatory power ($+0.1\%$ Accuracy, $+0.03$ AUC over $\lambda=0.8$). This confirms that a balanced combination, where accuracy drives the direction and entropy refines the confidence, yields the most robust decision boundary.
\end{itemize}

These findings confirm that DUA-D2C's dual-signal aggregation is strictly superior to relying on either metric individually.

\section{Limitations and Computational Efficiency Analysis} \label{lim}

We have demonstrated that the DUA-D2C framework significantly mitigates overfitting and enhances generalization across diverse domains. However, this performance gain involves a trade-off in training complexity. Dividing the training set into multiple subsets means the training process is effectively repeated $N$ times (where $N$ is the number of subsets), albeit on smaller data shards.

\subsection{Training Overhead}
\begin{figure}[ht!]
\centering
\includegraphics[width=\columnwidth]{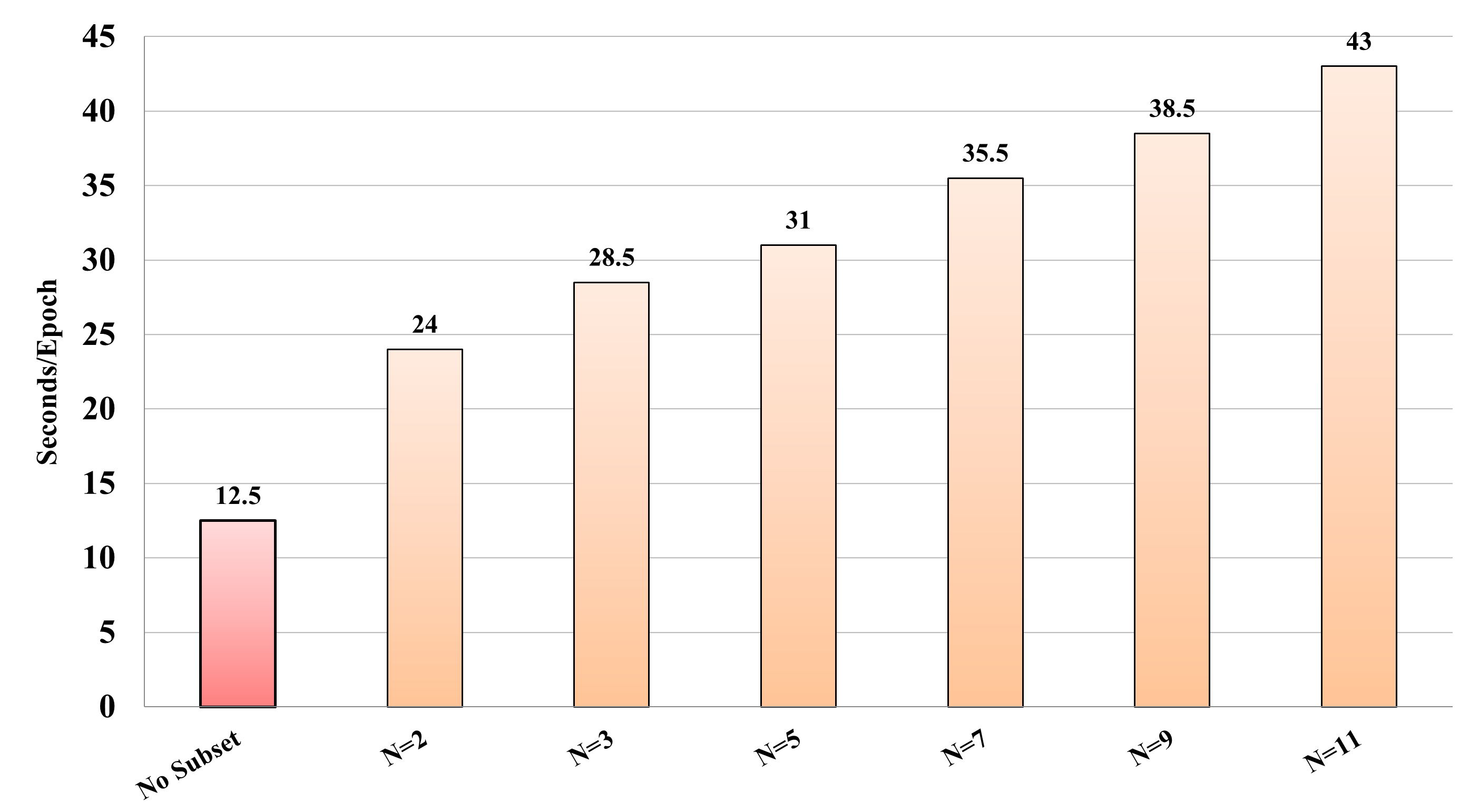}
\caption{Time elapsed per epoch for the traditional approach and different numbers of subsets, N, using the DUA-D2C method on the MNIST dataset.}
\label{fig:mnist-vary-n-bar}
\end{figure}

To quantify this cost, we analyzed the training time on the MNIST dataset using a single NVIDIA V100 GPU. As illustrated in Figure \ref{fig:mnist-vary-n-bar}, the monolithic traditional approach requires the least time per epoch (approx. 12.5 seconds). As $N$ increases, the total time to complete one global epoch (consisting of $N$ local trainings) increases. For instance, with $N=11$, the time per global epoch rose to 43 seconds.

Crucially, however, the increase is not proportional to $N$ (i.e., $11$ subsets did not take $11 \times$ longer) due to the reduced size of each shard. Furthermore, this limitation is largely an artifact of sequential simulation. \textbf{The inherent parallelism of the DUA-D2C framework lends itself perfectly to distributed computing.} In a real-world deployment, edge models can be trained simultaneously across multiple GPUs or nodes, potentially reducing the wall-clock training time to match that of a single epoch on the full dataset.

\subsection{Inference Efficiency vs. Standard Ensembles}
While training cost is higher, the primary advantage of DUA-D2C emerges during the deployment (inference) phase. Standard techniques for variance reduction, such as Bagging or Deep Ensembles, require maintaining $N$ distinct models and performing $N$ forward passes to aggregate predictions at runtime.

To quantify this advantage, we analyzed the computational complexity using the CNN architecture from our CIFAR-10 experiments (approx. 1.42 million parameters). Let $C_{fwd}$ be the FLOPs for one forward pass. Table \ref{tab:complexity} contrasts DUA-D2C against a standard ensemble of 5 models.

\begin{table}[ht]
    \renewcommand{\arraystretch}{1.2}
    \centering
    \caption{\label{tab:complexity} Inference Complexity Comparison: Traditional vs. Standard Ensemble ($N=5$) vs. DUA-D2C ($N=5$).}
    \begin{tabular}{|l|c|c|c|}
    \hline
    \textbf{Metric} & \textbf{Traditional} & \textbf{Ensemble ($N=5$)} & \textbf{DUA-D2C ($N=5$)} \\ \hline \hline
    \textbf{Architecture} & Single Model & 5 Distinct Models & Single Central Model \\ \hline
    \textbf{Inference FLOPs} & $1 \times C_{fwd}$ & $5 \times C_{fwd}$ & $\mathbf{1 \times C_{fwd}}$ \\ \hline
    \textbf{Parameter Count} & 1.42 M & 7.10 M & \textbf{1.42 M} \\ \hline
    \textbf{Memory Footprint} & $\sim 5.4$ MB & $\sim 27.0$ MB & $\mathbf{\sim 5.4}$ \textbf{MB} \\ \hline
    \textbf{Latency} & $T$ ms & $\approx 5T$ ms & $\mathbf{T}$ \textbf{ms} \\ \hline
    \end{tabular}
\end{table}

As shown in Table \ref{tab:complexity}, a standard ensemble inflates the memory footprint and inference latency by a factor of 5. In stark contrast, because DUA-D2C aggregates parameters rather than predictions, the final central model is architecturally identical to the baseline. Thus, DUA-D2C achieves the generalization benefits of an ensemble with \textbf{80\% less computational overhead} at inference time compared to a 5-model ensemble. This makes our method uniquely suitable for resource-constrained production environments where training time is a sunk cost but inference latency is critical.

\section{Conclusion} \label{conc}
Our study establishes a paradigm shift in deep learning regularization: treating data partitioning not as a constraint, but as a deliberate architectural strategy to mitigate overfitting. We demonstrated that while the foundational Divide2Conquer (D2C) framework reduces variance through isolation, our proposed \textbf{Dynamic Uncertainty-Aware Divide2Conquer (DUA-D2C)} perfects this technique by introducing an intelligent, quality-aware aggregation mechanism. By dynamically filtering out edge models that are uncertain or overfitted to local noise, DUA-D2C consistently produced significant performance gains across diverse domains.

Empirically, this advanced method outperformed both the traditional monolithic approach and static averaging schemes, showing significantly better resistance against overfitting as evidenced by delayed validation loss increases and smoother decision boundaries. These results substantiate our theoretical hypothesis that dynamic parameter fusion minimizes error variance more effectively than simple averaging. Moreover, we demonstrated that DUA-D2C is orthogonal to standard regularizers and incurs \textbf{zero additional computational cost} at inference time, offering a distinct advantage over standard ensembles. Thus, the proposed DUA-D2C framework establishes a robust, scalable solution for Big Data scenarios and opens the door for future research into intelligent distributed learning.

\begin{acks}
This work was supported by the Institute for Advanced Research (IAR), United International University, under grant UIU-IAR-02-2022-SE-06. The authors sincerely appreciate the funding and resources provided, which were essential to the successful execution of this research.
\end{acks}

\section*{Availability of Data and Codes}
The datasets analyzed during the study are publicly available. The source codes for this study are available on GitHub at \href{https://github.com/Saiful185/DUAD2C}{https://github.com/Saiful185/DUAD2C}.

\printbibliography
\appendix

\section{Appendix: Related Proofs and Derivations} \label{app1}
Derivation of $Var(X) = \E(X^2) - (\E(X))^2$ :
Let the random variable $X$ denote the estimate of $x$. The variance of estimating this random variable is defined as 
the average deviation of $X$ from the expected value of our
estimate, $\E(X)$, where the deviation is squared for symmetry of difference. So, the variance can be formulated as:
\begin{flalign}
    Var(X) &= \mathop{{}\mathbb{E}}((X - \mathop{{}\mathbb{E}}(X))^2)&&\nonumber\\
    &= \E(X^2 + (\E(X))^2 - 2X\E(X))&&\nonumber\\
    &\stackrel{\text{(a)}}{=} \E(X^2) + (\E(X))^2 - 2\E(X)\E(\E(X))&&\nonumber\\
     &= \E(X^2) - (\E(X))^2 &&\nonumber\\
    \label{proof1}
\end{flalign}
Here, $(a)$ is due to the linear nature of the expectation operator, and $\E(\E(X)) = \E(X)$, because $\E(X)$ is not a random variable.\\

Derivation of $CoV(XY) = \E(XY) - \E(X)\E(Y)$ : Let $X$ and $Y$ be estimations of two random variables $x$ and $y$ respectively.
The covariance of $X$ and $Y$ is defined as the expected value (or mean) of the product of their deviations from their individual expected values\cite{park2018fundamentals}. Hence,

\begin{flalign}
    Cov(X,Y) &= \E((X - \E(X)).(Y  - \E(Y))) &&\nonumber\\
    &= \E(XY - X\E(Y) - Y\E(X) + \E(X)\E(Y))&&\nonumber\\
    &\stackrel{\text{(b)}}{=}  \E(XY) - \E(X\E(Y)) &&\nonumber\\
    &\qquad- \E(Y\E(X)) + \E(\E(X)\E(Y))&&\nonumber\\
    &\stackrel{\text{(c)}}{=} \E(XY) -\E(X)\E(Y) &&\nonumber\\
    &\qquad- \E(Y)\E(X) + \E(X)\E(Y)&&\nonumber\\
    &= \E(XY) -\E(X)\E(Y)\nonumber&&\\
    \label{proof2}
\end{flalign}
Here, $(b)$ is because expectation is linear and $(c)$ is due to considering, $\E(cX) = c\E(X)$. Now if we consider two continuous random variables $X$ and $Y$ which are independent, i.e., $X\independent Y$, then by the definition of expectation we have:
\begin{flalign}
    \E(XY) &= \iint xy f(x,y) dxdy&&\nonumber\\
    &\stackrel{\independent}{=} \iint xy f(x)f(y) dx dy &&\nonumber\\
    &=\int yf(y) \int xf(x) dxdy &&\nonumber\\
    &\stackrel{\text{(d)}}{=}\int yf(y) \E(x)dy  &&\nonumber\\
    &= \E(X)\int yf(y) dy &&\nonumber\\
    &= \E(X)E(Y) \nonumber&&\\
    \label{proof3}
\end{flalign}

Here, $(d)$ is because $\E(X) = \int xf(x) dx$ by definition. From \ref{proof2} and \ref{proof3} we immediately have $Cov(X,Y) = 0$ when $X\independent Y$.\\

Derivation of $Var(\sum_{i=1}^k X_i)=\sum_{i=1}^k Var(X_i) + \sum_{1\leq i\neq j \leq k} Cov(X_i,X_j)$:\\ 
\\
Let us consider the random variables $X$ and $Y$ and arbitrary constants $a,b$ . Using \ref{proof1} we can write,

\begin{flalign}
    Var(aX + bY) &= \E((aX+bY)^2) - (\E(aX+bY))^2 &&\nonumber\\
    &= (a^2\E(X^2) +b^2\E(Y^2) +2ab\E(XY)) &&\nonumber\\
    &\qquad - (a\E(x)+b\E(Y))^2&&\nonumber\\
    &= (a^2\E(X^2) +b^2\E(Y^2)+2ab\E(XY))&&\nonumber\\
    &\qquad -a^2(\E(X))^2- b^2(\E(Y))^2 &&\nonumber\\
    &\qquad - 2ab\E(X)\E(Y) &&\nonumber\\
    &\stackrel{\text{(e)}}{=} a^2Var(X) + b^2Var(Y) + 2abCov(X,Y) &&\nonumber\\
    \label{proof4}
\end{flalign}

Here, $(e)$ follows directly from \ref{proof1} and \ref{proof2}. So Using \ref{proof3} and \ref{proof4} and generalizing for $k$ random variables we can say:\\

\begin{flalign}
    Var(\sum_{i=1}^k X_i) &= \sum_{i=1}^k Var(X_i) + \sum_{1\leq i\neq j \leq k} Cov(X_i,X_j)
\end{flalign}

\end{document}